\documentclass[runningheads]{llncs}
\usepackage{graphicx}

\usepackage{tikz}
\usepackage{comment}
\usepackage{amsmath,amssymb} %
\usepackage{color}

\usepackage[accsupp]{axessibility}  %

\usepackage{epsfig}
\usepackage{graphicx}
\usepackage{amsmath}
\usepackage{amssymb}
\usepackage{bm}

\usepackage{bbm}
\usepackage{inconsolata}
\usepackage{subcaption}
\usepackage{multirow}
\usepackage{booktabs}   %
\usepackage{makecell}   %
\usepackage{changepage} %
\usepackage{comment}

\usepackage{wrapfig}

\usepackage[pagebackref,breaklinks,colorlinks,breaklinks=true]{hyperref}
\usepackage{breakcites}

\usepackage{mathtools}  %

\usepackage{pifont}
\newcommand{\cmark}{\ding{51}}
\newcommand{\xmark}{\ding{55}}

\usepackage{tabularx}
\newcolumntype{R}{>{\raggedleft\arraybackslash}X}

\usepackage{longtable}

\usepackage{inconsolata}

\makeatletter
\usepackage{xspace}
\def\@onedot{\ifx\@let@token.\else.\null\fi\xspace}
\DeclareRobustCommand\onedot{\futurelet\@let@token\@onedot}

\def\eg{\emph{e.g}\onedot} 
\def\ie{\emph{i.e}\onedot} \def\Ie{\emph{I.e}\onedot}
 
 \def\vs{\emph{vs}\onedot}

\newcommand{\figref}[1]{Fig\onedot~\ref{#1}}
\newcommand{\equref}[1]{Eq\onedot~\eqref{#1}}
\newcommand{\secref}[1]{Sec\onedot~\ref{#1}}
\newcommand{\tabref}[1]{Tab\onedot~\ref{#1}}

\usepackage{csquotes}

\usepackage[normalem]{ulem}

\usepackage{multirow}

\usepackage{listings}

\usepackage{adjustbox}

\usepackage{fancybox, graphicx}

\usepackage{threeparttable}

\newcommand{\stylegan}{StyleGANv2}
\newcommand{\afhq}{AFHQv2}

\usepackage{color, colortbl}
\definecolor{Gray}{gray}{0.9}
\newcolumntype{g}{>{\columncolor{Gray}}r}
\definecolor{highlightRowColor}{rgb}{0.95, 0.95, 1}

\usepackage{array}
\usepackage{booktabs}
\usepackage[font=small]{caption}
\makeatletter
\g@addto@macro{\endtabular}{\gdef\rowfonttype{}}%
\makeatother
\newcommand{\rowfonttype}{}%
\newcolumntype{L}{>{\rowfonttype\strut}l}

\newcommand*{\belowrulesepcolor}[1]{%
  \noalign{%
    \kern-\belowrulesep 
    \begingroup 
      \color{#1}%
      \hrule height\belowrulesep 
    \endgroup 
  }%
} 
\newcommand*{\aboverulesepcolor}[1]{%
  \noalign{%
    \begingroup 
      \color{#1}%
      \hrule height\aboverulesep 
    \endgroup 
    \kern-\aboverulesep 
  }%
} 

\newcommand{\beginsupplement}{
    \setcounter{table}{0}
    \renewcommand{\thetable}{S\arabic{table}}%
    \setcounter{figure}{0}
    \renewcommand{\thefigure}{S\arabic{figure}}%
    \setcounter{equation}{0}
    \renewcommand{\theequation}{S\arabic{equation}}
}

\usepackage[resetlabels]{multibib}
\newcites{supp}{References}

\usepackage[capitalize]{cleveref}
\crefname{section}{Sec.}{Secs.}
\Crefname{section}{Section}{Sections}
\Crefname{table}{Table}{Tables}
\crefname{table}{Tab.}{Tabs.}
\crefformat{footnote}{#2\footnotemark[#1]#3}

\begin{document}
\pagestyle{headings}
\mainmatter
\def\ECCVSubNumber{4610}  %

\title{Generative Multiplane Images:\\Making a 2D GAN 3D-Aware} %

\titlerunning{Generative Multiplane Images}

\authorrunning{X.~Zhao, F.~Ma, D.~G\"{u}era, Z.~Ren, A.~G.~Schwing, A.~Colburn}

\author{
    Xiaoming Zhao\inst{1,2}\thanks{Work done as part of an internship at Apple.}, Fangchang Ma\inst{1}, David G\"{u}era\inst{1}, Zhile Ren\inst{1}, \\
    Alexander G. Schwing\inst{2}\index{Schwing, Alexander G.}, Alex Colburn\inst{1}
}
\institute{
Apple \and 
University of Illinois, Urbana-Champaign\\
\url{https://github.com/apple/ml-gmpi}
}
\maketitle

\begin{abstract}
What is really needed to make an existing 2D GAN 3D-aware? To answer this question, we modify a classical GAN,~\ie,~\stylegan, as little as possible. We find that only two modifications are absolutely necessary: 1) a multiplane image style generator branch which produces a set of alpha maps conditioned on their depth; 2) a pose-conditioned discriminator. We refer to the generated output as a `generative multiplane image' (GMPI)
and emphasize that its renderings are not only high-quality but also guaranteed to be view-consistent, which makes GMPIs different from many prior works. Importantly, the number of alpha maps can be dynamically adjusted and can differ between training and inference, alleviating memory concerns and enabling fast training of GMPIs in less than half a day at a resolution of $1024^2$. Our findings are consistent across  three challenging and common high-resolution datasets, including FFHQ, \afhq~and MetFaces. 
\end{abstract}

\section{Introduction}

\begin{figure}[!t]
    \centering
    \begin{tabular}{cc}
    \raisebox{1.5cm}[0pt][0pt]{\rotatebox[origin=c]{90}{FFHQ $1024^2$}}&\includegraphics[width=0.9\textwidth]{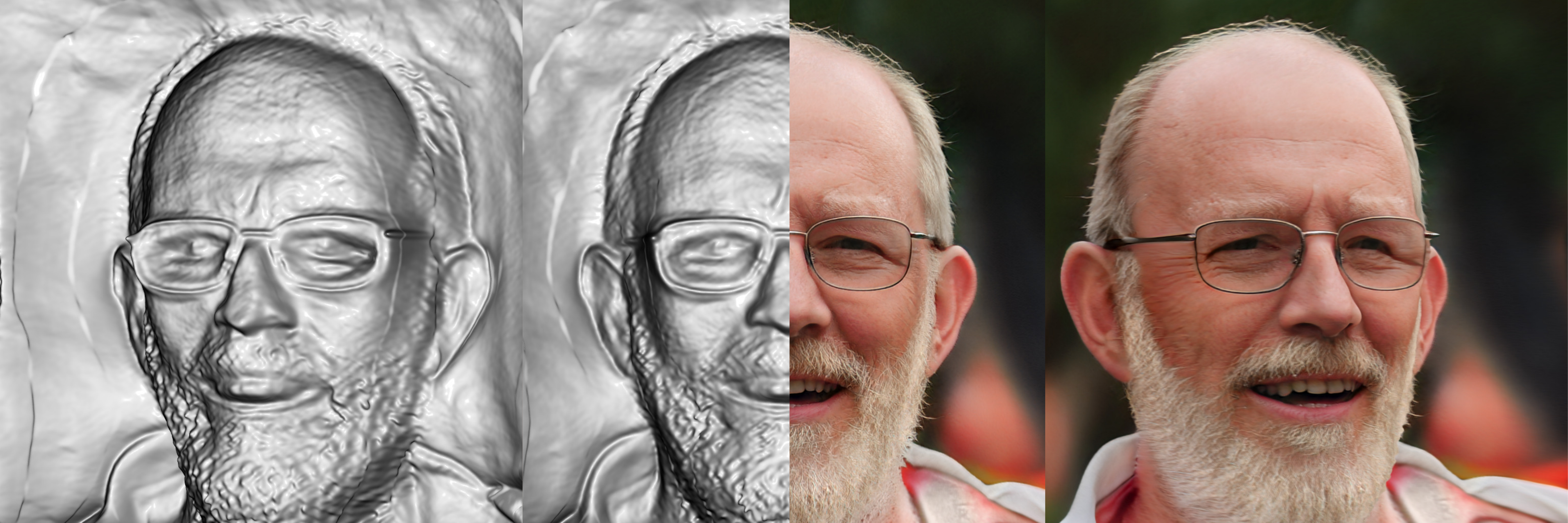}\\
    \raisebox{1.5cm}[0pt][0pt]{\rotatebox[origin=c]{90}{\afhq~$512^2$}}&\includegraphics[width=0.9\textwidth]{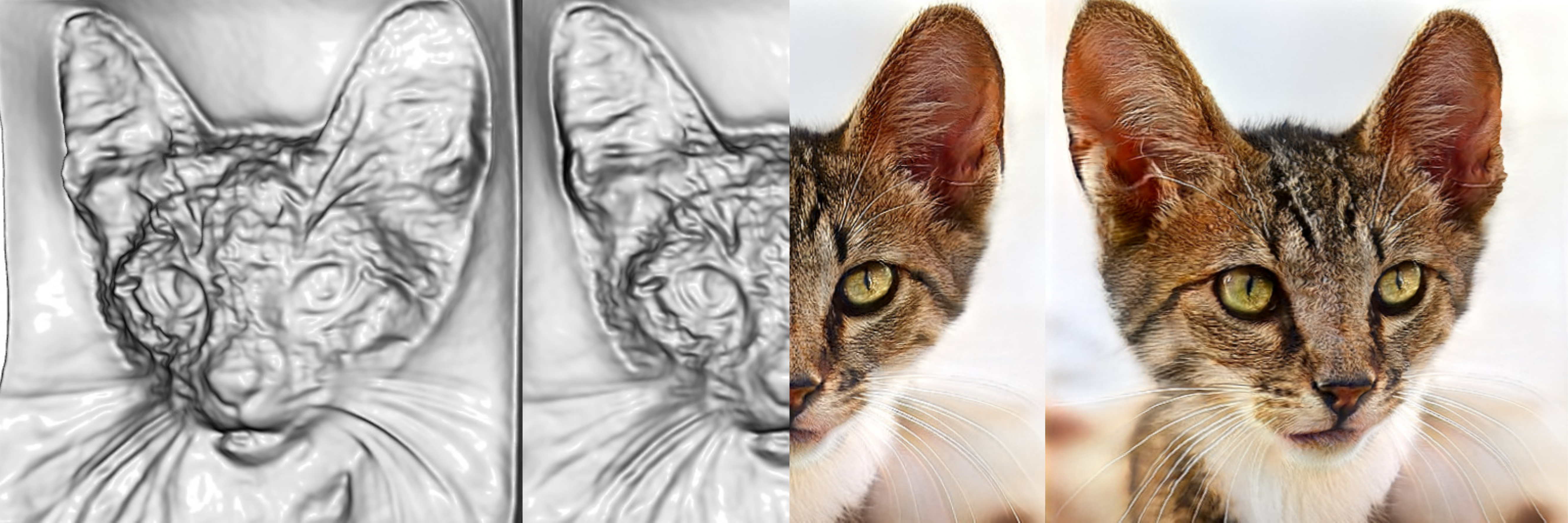}\\
    \raisebox{1.5cm}[0pt][0pt]{\rotatebox[origin=c]{90}{MetFaces $1024^2$}}&\includegraphics[width=0.9\textwidth]{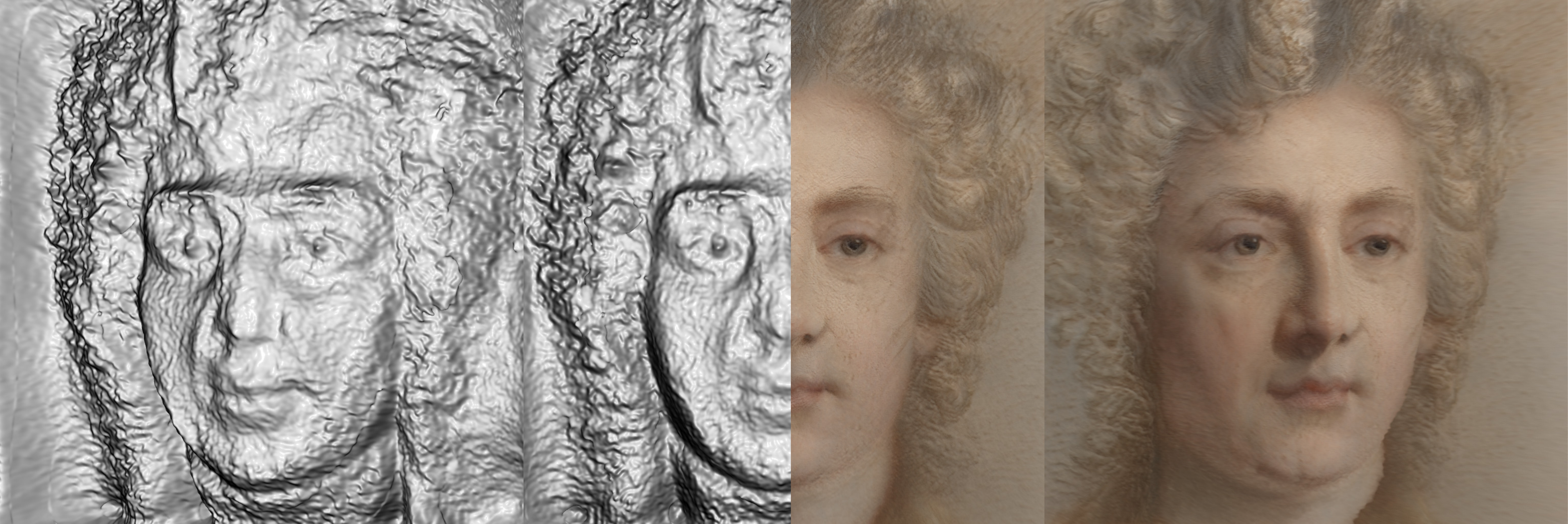}
    \end{tabular}
    \caption{Making a 2D GAN 3D-aware with a minimal set of changes: on three datasets (FFHQ, \afhq~and MetFaces) we observe that it suffices to augment \stylegan\ with an additional branch that generates alpha maps conditioned on  depths, and to modify training by using a discriminator conditioned on pose.}
    \label{fig:teaser}
\end{figure}

Generative adversarial networks (GANs)~\cite{Goodfellow2014GenerativeAN} have been remarkably successful at sampling novel images which look `similar' to those from a given training dataset. 
Notably, impressive advances have been reported in recent years which improve  quality and resolution of the generated images. 
Most of these advances focus on the setting where the output space of the generator and the given dataset are identical and, often, these outputs are either images or occasionally 3D volumes.

The latest literature, however, has focused on generating novel outputs that differ from the available training data. This includes methods that generate 3D geometry and the corresponding texture for one class of objects,~\eg,~faces, while the given dataset only contains widely available single-view images~\cite{SchwarzNEURIPS2020,Niemeyer2021GIRAFFERS,ChanCVPR20201,OrEl2021StyleSDFH3,Deng2021GRAMGR,ChanARXIV2021,ZhouARXIV2021,xu2021volumegan}. No multi-view images or 3D geometry  are used to supervise the training of these 3D-aware GANs. To learn the 3D geometry from such a limited form of supervision, prior work typically combines 3D-aware inductive biases such as a 3D voxel grid or an implicit representation with a rendering engine~\cite{SchwarzNEURIPS2020,Niemeyer2021GIRAFFERS,ChanCVPR20201,OrEl2021StyleSDFH3,Deng2021GRAMGR,ChanARXIV2021,ZhouARXIV2021,xu2021volumegan}.

Nevertheless, improving the quality of the results of these methods remains challenging: 
3D-aware inductive biases are often memory-intensive explicit or implicit 3D volumes, and/or rendering is often computationally demanding,~\eg,~involving a two-pass importance sampling in a 3D volume and a subsequent decoding of the obtained features. 
Moreover, lessons learned from 2D GANs are often not directly transferable because the generator output or even its entire structure has to be adjusted.  This poses the question:
\begin{displayquote}
\textit{``What is really needed to make an existing 2D GAN 3D-aware?''}
\end{displayquote}

To answer this question, we aim to modify an existing 2D GAN as little as possible. Further, we aim for an efficient inference and training procedure. As a starting point, we chose the widely used \stylegan~\cite{Karras2020AnalyzingAI}, which has the added benefit that many training checkpoints are publicly available.

More specifically, we develop a new generator branch for \stylegan\ which yields  a set of fronto-parallel alpha maps, in spirit similar to multiplane images (MPIs)~\cite{ZhouTOG2018}. As far as we know, we are the first to demonstrate that MPIs can be used as a scene representation for \textit{unconditional} 3D-aware generative models.
This new alpha branch is trained from scratch while the regular \stylegan\ generator and  discriminator are simultaneously fine-tuned. Combining the generated alpha maps with the single standard image output of \stylegan\ in an end-to-end differentiable multiplane style rendering, we obtain a 3D-aware generation from different views while \emph{guaranteeing view-consistency}. Although alpha maps have limited ability to handle occlusions, rendering is very efficient. Moreover, the number of alpha maps can be dynamically adjusted and can even differ between training and inference, alleviating memory concerns. We refer to the generated output of this method as a `generative multiplane image' (GMPI). 

To obtain alpha maps that exhibit an expected 3D structure, we find that only two adjustments  of \stylegan~are really necessary: (a) the alpha map prediction of any plane in the MPI has to be conditioned on the plane's depth or a learnable token; (b) the discriminator has to be conditioned on camera poses. While these two adjustments seem intuitive in hindsight, it is still surprising that an alpha map with planes conditioned on their depth and use of camera pose information in the discriminator are sufficient inductive biases for 3D-awareness. 

An additional inductive bias that improves the alpha maps is a 3D rendering that incorporates shading. Albeit useful, we didn't find this inductive bias to be  necessary to obtain 3D-awareness.
Moreover, we find that metrics designed for classic 2D GAN evaluation,~\eg,~the Fr\'echet Inception Distance (FID)~\cite{Heusel2017GANsTB} and the Kernel Inception Distance (KID)~\cite{Binkowski2018DemystifyingMG} may lead to misleading results %
since they do not consider geometry.

In summary, our contribution is two-fold:
1) We are the first to study an MPI-like 3D-aware generative model trained with standard single-view 2D image datasets;
2) We find that conditioning the alpha planes on depth or a learnable token and the discriminator on camera pose are sufficient to make a 2D GAN 3D-aware. Other information provides improvements but is not strictly necessary. %
We study the aforementioned ways to encode 3D-aware inductive biases on three high-resolution datasets: FFHQ~\cite{Karras2019ASG}, \afhq~\cite{choi2020starganv2}, and MetFaces~\cite{Karras2020TrainingGA}. Across all three datasets, as illustrated in \cref{fig:teaser}, our findings are consistent.

\section{Related Work and Background}

In the following, we briefly review recent advances in classical and neural scene rendering, as well as the generation of 2D and 3D data with generative models. We then discuss the generation of 3D data using only 2D supervision. We also provide a brief review of single image reconstruction techniques before we review \stylegan\ in greater detail.

\noindent\textbf{Scene Rendering.} Image-Based Rendering (IBR) is well-studied~\cite{ChenSIGGRAPH1993}. IBR 1) models a scene 
given a set of images; and 2) uses the scene model to render novel views. 
Methods can be grouped based on their use of scene geometry: explicit, implicit, or not using geometry.
Texture-mapping methods use explicit geometry, whereas layered depth images (LDIs)~\cite{ShadeSIGGRAPH1998}, flow-based~\cite{ChenSIGGRAPH1993}, lumigraphs~\cite{BuehlerSIGGRAPH2001}, and tensor-based methods~\cite{AvidanCVPR1997} use geometry implicitly. In contrast, light-field methods~\cite{LevoySIGGRAPH1996} do not rely on geometry.
Hybrid methods~\cite{DebevecSIGGRAPH1996} have also  been studied. 
More recently, neural representations have been used in IBR, for instance, neural radiance fields (NeRFs)~\cite{MildenhallECCV2020} and multiplane images (MPIs)~\cite{ZhouTOG2018,srinivasan2019pushing,tucker2020single,ghosh2021liveview,li2021mine}.
In common to both is the goal to extract from a given set of images a volumetric representation of the observed scene. 
The volumetric representation in MPIs is discrete and permits fast rendering, whereas NeRFs use a continuous spatial representation.

These works differ from the proposed method in two main aspects. 1) The proposed method is unconditionally generative,~\ie,~novel, never-before-seen scenes can be synthesized without requiring any color~\cite{tucker2020single}, depth, or semantic images~\cite{habtegebrial2020generative}. In contrast, IBR methods and related recent advances focus on reconstructing a scene representation from a set of images. 2) The proposed method uses a collection of `single-view images' from different scenes during training. In contrast, IBR methods use multiple viewpoints of a single scene for highly accurate reconstruction. %

\noindent\textbf{2D Generative Models.} Generative adversarial networks (GANs)~\cite{Goodfellow2014GenerativeAN} and variational auto-encoders (VAEs)~\cite{Kingma2014AutoEncodingVB} significantly advanced modeling of probability distributions. In the early days, GANs were notably difficult to train whereas VAEs often produced blurry images. However, in the last decade, theoretical and practical understanding of these methods has significantly improved. 
New loss functions and other techniques have been proposed~\cite{li2017mmd,gulrajani2017improved,kolouri2017sliced,deshpande2018generative,cully2017magan,mroueh2017mcgan,berthelot2017began,mroueh2017fisher,lin2017pacgan,heusel2017gans,salimans2018improving,Mescheder2018WhichTM,Arjovsky2017WassersteinG,Deshpande2019MaxSlicedWD,Aneja2020ACL,LiNIPS2017,SunNEURIPS2020} to improve the stability of GAN optimization  and to address mode-collapse, some theoretically founded
and others empirically motivated. 
We follow the architectural design and techniques in \stylegan~\cite{Karras2020AnalyzingAI}, including exponential moving average (EMA), gradient penalties~\cite{Mescheder2018WhichTM}, and minibatch standard deviation~\cite{Karras2018ProgressiveGO}.

\noindent\textbf{3D Generative Models from 2D Data.} 3D-aware image synthesis has gained attention recently.
Sparse volume representations are used to generate photorealistic images based on given geometry input~\cite{HaoICCV2021}.
Many early approaches use voxel-based representations~\cite{ZhuNEURIPS2018,WuNIPS2016,NguyenNEURIPS2020,NguyenICCV2019,HenzlerICCV2019,Gadelha3DV2017}, where scaling to higher resolutions is prohibitive due to the high memory footprint. 
Rendering at low-res followed by 2D CNN-based upsampling~\cite{NiemeyerCVPR2021} has been proposed as a workaround, but it leads to view inconsistency.
As an alternative, methods built on implicit functions,~\eg,~NeRF, have been proposed~\cite{ChanCVPR20201,SchwarzNEURIPS2020,Pan2021ASG}. However, their costly querying and sampling operations limit training efficiency and image resolutions.
To generate high-resolution images, concurrently, EG3D~\cite{ChanARXIV2021}, StyleNeRF~\cite{GuARXIV2021}, CIPS-3D~\cite{ZhouARXIV2021}, VolumeGAN~\cite{xu2021volumegan}, and StyleSDF~\cite{OrEl2021StyleSDFH3} have been developed.
Our work differs primarily in the choice of scene representation: EG3D uses a hybrid tri-plane representation while the others follow a NeRF-style implicit representation. In contrast, we study an MPI-like representation.
In our experience, MPIs provide extremely fast rendering speed without incurring quality degradation.
Most related  to our work  are GRAM~\cite{Deng2021GRAMGR} and LiftedGAN~\cite{Shi2021Lifting2S}.
GRAM uses a NeRF-style scene representation and learns scene-specific isosurfaces. Queries of RGB and density happen on those isosurfaces.
Although isosurfaces are conceptually similar to MPIs, the queries of this NeRF-style representation are expensive, limiting the image synthesis to low resolution at $256 \times 256$.
LiftedGAN reconstructs the geometry of an image by distilling intermediate representations from a fixed \stylegan~to a separate 3D generator which produces a depth map and a transformation map in addition to an image. Different from our proposed approach, because of the transformation map, LiftedGAN is not strictly view-consistent. Moreover, the use of a single depth-map is less flexible.

\noindent\textbf{\stylegan~Revisited.} Since our method is built on \stylegan, we provide some background on its architecture. \stylegan~generates a square 2D image $C \in \mathbb{R}^{H\times H\times 3}$ by upsampling and accumulating intermediate GAN generation results from various resolutions. Formally, the image $C \triangleq C^H$ is obtained via
\begin{align}
    C^{h} = 
    \begin{cases}
        \tilde{C}^h + \texttt{UpSample}_{\frac{h}{2}\rightarrow h} (C^{\frac{h}{2}}),\; &\text{if } h \in {\cal R} \setminus \{ 4 \}, \\
        \tilde{C}^4, &\text{if } h = 4,
    \end{cases} \label{eq: stylegan C}
\end{align}
where $C^h \in \mathbb{R}^{h\times h \times 3}$ is the GAN image generation at resolution $h$ and $\tilde{C}^h$ is the generated residual at the same resolution. ${\cal R} = \{4, 8, \dots, H\}$ is the set of available resolutions whose values are powers of 2. $\texttt{UpSample}_{\frac{h}{2}\rightarrow h}$ refers to the operation that upsamples from $\mathbb{R}^{\frac{h}{2}\times \frac{h}{2} \times 3}$ to $\mathbb{R}^{h\times h \times 3}$. The residual generation $\tilde{C}^h$ at resolution $h$ is generated with a single convolutional layer $f_\texttt{ToRGB}^h$,~\ie,
\begin{align}
    \tilde{C}^h = f_\texttt{ToRGB}^h (\mathcal{F}^h),\label{eq: stylegan C^h}
\end{align}
where $\mathcal{F}^h \in \mathbb{R}^{h \times h \times \texttt{dim}_h}$ is the intermediate GAN feature representation at resolution $h$ with $\texttt{dim}_h$ channels. These intermediate GAN feature representations at all resolutions ${\cal R}$ are computed with a synthesis network $f_\texttt{Syn}$,~\ie, 
\begin{align}
    \{\mathcal{F}^h : h \in {\cal R}\} = f_\texttt{Syn} (\bm{\omega}), %
    \label{eq: stylegan w}
\end{align}
 where $\bm{\omega}$ is the style embedding. $\bm{\omega}$  is computed via the mapping network $f_\texttt{Mapping}$ which operates on the latent variable $\bm{z}$,~\ie, $\bm{\omega} = f_\texttt{Mapping} (\bm{z})$.

\section{Generative Multiplane Images (GMPI)}

Our goal is to adjust an existing GAN such that it generates images which are 3D-aware and view-consistent,~\ie,~the image $I_{v_\texttt{tgt}}$ can illustrate the \emph{exactly identical} generated object from different camera poses $v_\texttt{tgt}$. 
In order to achieve 3D-awareness and guaranteed view-consistency, different from existing prior work,  we aim to augment an existing generative adversarial network, in our case \stylegan, as little as possible. For this, we modify the classical generator by adding an `alpha branch' and incorporate a simple and efficient alpha composition rendering. Specifically, the `alpha branch' produces alpha maps of a multiplane image while the alpha composition rendering step transforms generated alpha maps and generated image into the target view $I_{v_\texttt{tgt}}$ given a user-specified pose $v_\texttt{tgt}$. We refer to the output of the generator and the renderer as a `generative multiplane image' (GMPI). To achieve 3D-awareness, we also find the pose conditioning of the discriminator to be absolutely necessary. Moreover,  additional miscellaneous adjustments like the use of shading help to improve results. We first  discuss the generator, specifically our alpha branch (\cref{sec:mib}) and our rendering (\cref{sec:r}). Subsequently we detail the discriminator pose conditioning (\cref{sec:dpc}) and the miscellaneous adjustments (\cref{sec:ma}).

\begin{figure}[!t]
    \centering
    \includegraphics[width=0.95\textwidth]{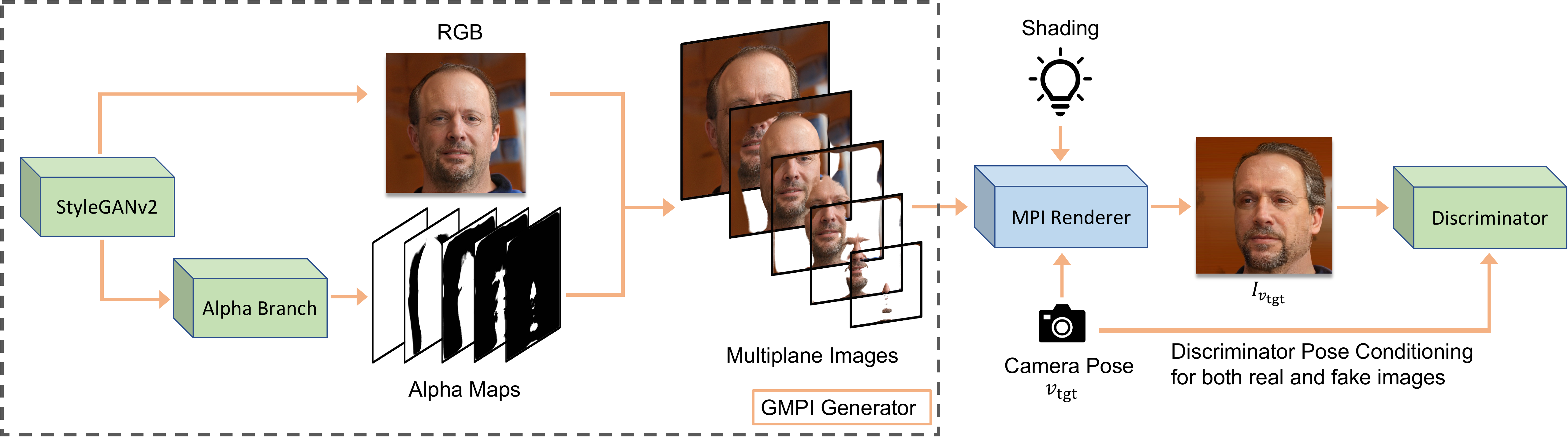}
    \caption{
    Overview of the developed 3D-aware generative multiplane images (GMPI).
    We find that two components are necessary to make a 2D \stylegan~3D-aware: accompanying \stylegan~with an MPI branch (\secref{sec:mib}) and conditioning the discriminator  on pose during training (\secref{sec:dpc}).
    We find adding shading during training (\secref{sec:ma}) improves the generated geometry.
    Please also refer to~\secref{sec:r} for  details about the employed MPI rendering.
    Green blocks denote trainable components.
    }
    \label{fig:overview}
\end{figure}

\subsection{Overview}\label{sec: overview}
An overview of our method to generate GMPIs  is shown in \cref{fig:overview}. Its generator and the subsequent alpha-composition renderer produce an image $I_{v_\texttt{tgt}}$ illustrating the generated object from a user-specified pose $v_\texttt{tgt}$. Images produced for different poses are guaranteed to be view-consistent. 
The generator and rendering achieve 3D-awareness and guaranteed view consistency in two steps.
First, a novel `alpha branch'  uses intermediate representations  to produce a multiplane image representation ${\cal M}$ which contains alpha maps at various depths in addition to a single image. Importantly, to obtain proper 3D-awareness we find that it is necessary to condition alpha maps on their depth. We discuss the architecture of this alpha branch and its use of the plane depth in \cref{sec:mib}. Second, to guarantee view consistency, we employ a rendering step (\secref{sec:r}). It converts the  representation ${\cal M}$ obtained from the alpha branch  into the view $I_{v_\texttt{tgt}}$, which shows the generated object from a user-specified pose $v_\texttt{tgt}$. %
During training, the generated image $I_{v_\texttt{tgt}}$ is  compared to real images from a single-view dataset via a pose conditioned discriminator. We discuss the discriminator details in \cref{sec:dpc}. Finally, in \cref{sec:ma}, we highlight some miscellaneous adjustments which we found to help improve 3D-awareness while not being strictly necessary.

\begin{figure}[!t]
    \centering
    \includegraphics[width=0.75\textwidth]{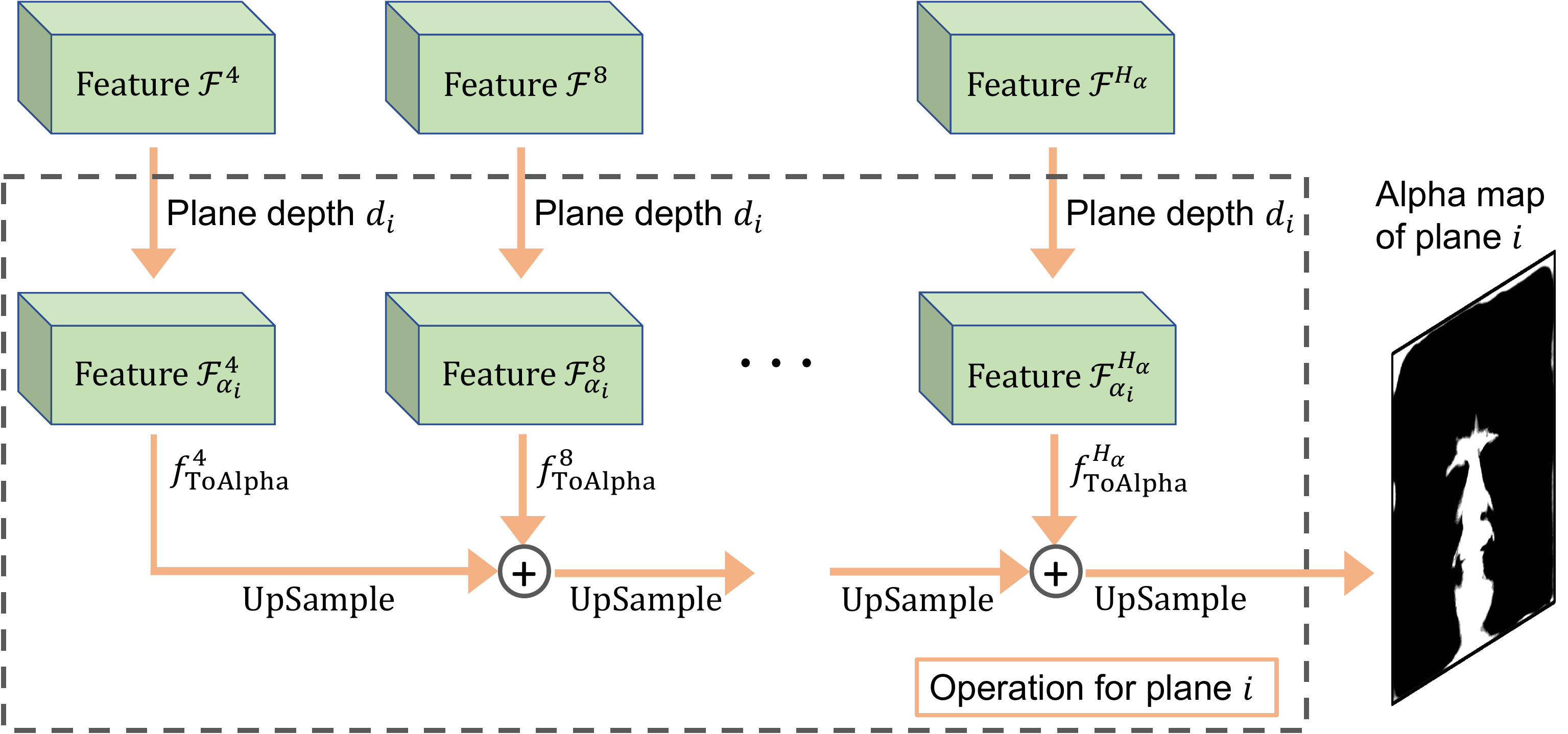}
    \caption{
    The alpha branch proposed in \secref{sec:mib}.
    Here we show the generation of the  alpha map of plane $i$. At each intermediate resolution $h \in {\cal R}_\alpha = \{4, 8, \dots, H_\alpha\}$, we utilize the plane's depth $d_i$ to transform the feature from ${\cal F}^h$ (\equref{eq: stylegan w}) to $\mathcal{F}_{\alpha_i}^h$ (\equref{eq: F_alpha}). %
    The final alpha map $\alpha_i$ is obtained by accumulating  all intermediate results, which are generated by the single convolutional layer $f_\texttt{ToAlpha}^h$.
    }
    \label{fig:mib}
\end{figure}

\subsection{GMPI: {\stylegan} with an Alpha Branch}
\label{sec:mib}
Given an input latent code $\bm{z}$, our goal is to synthesize a multiplane image inspired representation $\mathcal{M}$ from which realistic and view-consistent 2D images can be rendered at different viewing angles. A classical multiplane image refers to a set of tuples $(C_i, \alpha_i, d_i)$ for $L$ fronto-parallel planes $ i \in \{1, \dots, L\}$. %
Within each tuple, $C_i \in \mathbb{R}^{H\times H\times 3}$ denotes the color texture for the $i^\text{th}$ plane and is assumed to be a square image of size $H\times H$, where $H$ is independent of the plane index $i$. Similarly, $\alpha_i \in [0,1]^{H\times H\times 1}$ and $d_i\in\mathbb{R}$ denote the alpha map and the depth of the corresponding plane,~\ie,~its distance from a camera. $i=1$ and $i=L$ denote the planes closest and farthest from the camera. %

As a significant simplification, we choose to reuse the same color texture image $C_i \triangleq C \,\forall\, i$ across all planes. $C$ is synthesized by the original \stylegan\ structure as specified in \equref{eq: stylegan C}. %
Consequently, the task of the generator $f_\texttt{G}$ has been reduced from predicting an entire multiplane image representation  to predicting a single color image and a set of alpha planes, \ie, 
\begin{align}
    \mathcal{M} \triangleq \left\{C, \{\alpha_1, \alpha_2, \dots, \alpha_L\}\right\} = f_\texttt{G}(\bm{z}, \{d_1, d_2, \dots, d_L\}).
    \label{eq: mpi define}
\end{align}

For this, we propose a simple modification of the original \stylegan\ network by 
adding an additional alpha branch, as illustrated in \cref{fig:mib}. For consistency we follow \stylegan's design: the alpha map $\alpha_i \triangleq \alpha_i^H$ of plane $i$ is  obtained by upsampling and accumulating alpha maps at different resolutions. Notably, we do not  generate all the way up to the highest resolution $H$, but instead use the final upsampling step
\begin{align}
    \alpha_i \triangleq \alpha_i^H = \texttt{UpSample}_{H_\alpha \rightarrow H} (\alpha_i^{H_\alpha}),\label{eq: alpha H_a}
\end{align}
if $H_\alpha < H$. Here $H_\alpha \leq H$ refers to a possibly lower resolution. We will explain the reason for this design below.
Following~\equref{eq: stylegan C}, we have
\begin{align}
    \alpha_i^h = 
    \begin{cases}
        \tilde{\alpha}_i^h + \texttt{UpSample}_{\frac{h}{2}\rightarrow h} (\alpha_i^{\frac{h}{2}}),\; &\text{if } h \in {\cal R_\alpha} \setminus \{ 4 \}, \\
        \tilde{\alpha}_i^4, &\text{if } h = 4.
    \end{cases} \label{eq: stylegan alpha}
\end{align}
Here, $\alpha_i^h \in [0,1]^{h \times h \times 1}$ and $\tilde{\alpha}_i^h$ are the alpha map and the residual at resolution $h$ respectively. ${\cal R_\alpha} \triangleq \{4, 8, \dots, H_\alpha \} \subseteq \cal R$ denotes the set of alpha maps' intermediate resolutions.
Inspired by \equref{eq: stylegan C^h}, $\tilde{\alpha}_i^h$ is generated from an intermediate feature representation $\mathcal{F}_{\alpha_i}^h$ through a single convolutional layer $f^h_\texttt{ToAlpha}$:
\begin{align}
    \tilde{\alpha}_i^h = f_\texttt{ToAlpha}^h(\mathcal{F}_{\alpha_i}^h).\label{eq: to alpha}
\end{align}
Note, $f_\texttt{ToAlpha}^h$ is shared across all planes $i$, while the input feature $\mathcal{F}_{\alpha_i}^h$ is plane-specific. Inspired by AdaIn~\cite{Huang2017ArbitraryST}, we construct this plane-aware feature as follows:
\begin{align}
    \mathcal{F}_{\alpha_i}^h = \frac{\mathcal{F}^h - \mu(\mathcal{F}^h)}{\sigma(\mathcal{F}^h)} + f_\texttt{Embed}(d_i, \bm{\omega}), \label{eq: F_alpha}
\end{align}
where $\mu(\mathcal{F}^h) \in \mathbb{R}^{\texttt{dim}_h}$ and $\sigma(\mathcal{F}^h) \in \mathbb{R}^{\texttt{dim}_h}$ are mean and standard deviation of the feature $\mathcal{F}^h \in \mathbb{R}^{h \times h \times \texttt{dim}_h}$ from \equref{eq: stylegan w}. Meanwhile, $f_\texttt{Embed}$ uses the depth $d_i$ of plane $i$ and the style embedding $\bm{\omega}$ from \equref{eq: stylegan w} to compute a plane-specific embedding in the space of $\mathbb{R}^{\texttt{dim}_h}$.
Note, our design of $\mathcal{F}_{\alpha_i}^h$  disentangles alpha map generation from a pre-determined  number of planes as $f_\texttt{ToAlpha}^h$  operates on each plane individually. This provides the ability to use an arbitrary number of planes which helps to reduce artifacts during inference as we will show later.

Note,  the plane specific feature $\mathcal{F}_{\alpha_i}^h \forall i$  will in total occupy $L$ times the memory used by the feature $\mathcal{F}^h$. This might be prohibitive if we were to generate these intermediate features up to a resolution of $H$. Therefore, we only use this feature until some lower resolution $H_\alpha$ in~\equref{eq: alpha H_a}. We will show later that this design choice works well on real-world data.

\subsection{Differentiable Rendering in GMPI}
\label{sec:r}
We obtain the desired image $I_{v_\texttt{tgt}}$ which illustrates the generated MPI representation 
${\cal M} = \left\{C, \{\alpha_1, \dots, \alpha_L\}\right\}$ 
from the user-specified target view $v_\texttt{tgt}$ with an MPI renderer in two steps: 1) a warping step transforms the  representation $\mathcal{M}$ from its canonical pose $v_\texttt{cano}$ to the target pose $v_\texttt{tgt}$; 2) a compositing step combines the planes into the desired image $I_{v_\texttt{tgt}}$. Importantly, both steps entail easy computations which are end-to-end differentiable such that they can be included into any generator.
Please see Appendix~\ref{supp sec: gmpi render}  for details.

\subsection{Discriminator Pose Conditioning}
\label{sec:dpc}
For the discriminator $f_\texttt{D}$  to encourage 3D-awareness of the generator, we find the conditioning of the discriminator on camera poses to be essential. Formally, inspired by Miyato and Koyama~\cite{Miyato2018cGANsWP}, the final prediction of the discriminator is 
\begin{align}
    \log P(y = \texttt{real}\vert I_{v_\texttt{tgt}}, v_\texttt{tgt}) \propto \texttt{Normalize} (f_\texttt{Embed}^\texttt{D}(v_\texttt{tgt})) \cdot f_\texttt{D} (I_{v_\texttt{tgt}})^\top, \label{eq: dpc}
\end{align}
where $P (y = \texttt{real} \vert I_{v_\texttt{tgt}},  v_\texttt{tgt})$ denotes the probability that image $I_{v_\texttt{tgt}}$ from the camera pose $ v_\texttt{tgt}$ is real.
$v_\texttt{tgt} \in \mathbb{R}^{16}$ denotes the flattened extrinsic matrix of the camera pose.
$f_\texttt{Embed}^\texttt{D}: \mathbb{R}^{16} \mapsto \mathbb{R}^{16}$ denotes an embedding function, while $\texttt{Normalize}(\cdot)$ results in a zero-mean, unit-variance embedding.
$f_\texttt{D}(\cdot) \in \mathbb{R}^{16}$ denotes the feature computed by the discriminator \texttt{D}.
For real images of faces,~\eg,~humans and cats, $v_\texttt{tgt}$ can be estimated via  off-the-shelf tools~\cite{Deng2019Accurate3F,CatHipsterizer}.\footnote{Concurrently, EG3D~\cite{ChanARXIV2021} also finds that  pose conditioning of the discriminator is required for their tri-plane representation to produce 3D-aware results, corroborating that this form of inductive bias is indeed necessary.}

\subsection{Miscellaneous Adjustment: Shading-guided Training}
\label{sec:ma}

Inspired by~\cite{Pan2021ASG}, we incorporate shading into the rendering process introduced in~\secref{sec:r}. Conceptually, shading will amplify artifacts of the generated geometry that might be hidden by texture, encouraging the alpha branch to produce better results.
To achieve this, we adjust the  RGB component $C\in \mathbb{R}^{H \times W \times 3}$ via
\begin{align}
    \widehat{C} = C \cdot \left( k_a + k_d \: \bm{l} \cdot \texttt{N} \left( D_{v_\texttt{cano}} \right)  \right),\label{eq: shaing}
\end{align}
where $k_a$ and $k_d$ are coefficients for ambient and diffuse shading, $\bm{l} \in \mathbb{R}^3$ indicates  lighting  direction, and $\texttt{N} \left(D_{v_\texttt{cano}}\right) \in \mathbb{R}^{H\times W \times 3}$ denotes the normal map computed from the depth map $D_{v_\texttt{cano}}$ (obtained by using the canonical alpha maps $\alpha_i$, see~\equref{eq: mpi over depth} in Appendix~\ref{supp sec: gmpi render}). 
We find shading to slightly improve results while not being  required for 3D-awareness. 
Implementation details are in Appendix~\ref{supp sec: implement}.

\subsection{Training}\label{sec: training}

\noindent\textbf{Model structure.} The trainable components of GMPI are $f_\texttt{ToRGB}^h$ (\equref{eq: stylegan C^h}), $f_\texttt{Syn}$ and $f_\texttt{Mapping}$ (\equref{eq: stylegan w}), $f_\texttt{ToAlpha}^h$ (\equref{eq: to alpha}), $f_\texttt{Embed}$ (\equref{eq: F_alpha}), and $f_\texttt{D}$ and $f_\texttt{Embed}^\texttt{D}$ (\equref{eq: dpc}).
Please check the appendix for  implementation details.
Our alpha maps (\equref{eq: mpi define}) are equally placed in the disparity (inverse depth) space during training and we set $H_\alpha = 256$ (\equref{eq: alpha H_a}).

\noindent\textbf{Initialization.} For any training, we initialize weights from the officially-released \stylegan~checkpoints.\footnote{\label{footnote: stylegan repo}\url{https://github.com/NVlabs/stylegan2-ada-pytorch}} This enables a much faster training as we will show.

\noindent\textbf{Loss.}  We use $\theta$ to subsume all trainable parameters. Our training loss consists of a non-saturating GAN loss with R1 penalties~\cite{Mescheder2018WhichTM}, \ie, 
\begin{align}
    \mathcal{L}_\theta &= \mathbb{E}_{I_{v_\texttt{tgt}}, v_\texttt{tgt}} \left[ f (\log P(y = \texttt{real}\vert I_{v_\texttt{tgt}}, v_\texttt{tgt}))  \right] \nonumber \\
    &+ \mathbb{E}_{I, v_\texttt{tgt}} \left[ f (\log P(y = \texttt{real}\vert I, v_\texttt{tgt})) + \lambda \vert \nabla_I \log P(y = \texttt{real}\vert I, v_\texttt{tgt}) \vert^2  \right].
\end{align}
Here $f(x) \!=\! - \log (1 \!+\! \exp(-x))$,  $\lambda \!=\! 10.0$ in all studies, $I$ refers to real images and $v_\texttt{tgt}$ denotes the corresponding observer's pose information.

\section{Experiments}
\label{sec:exp}
We analyze GMPI on three datasets (FFHQ, \afhq~and MetFaces) and across a variety of resolutions. We first provide details regarding the three datasets before discussing evaluation metrics and quantitative as well as qualitative results.

\subsection{Datasets}
\noindent\textbf{FFHQ.} The FFHQ dataset~\cite{Karras2019ASG} consists of 70,000 high-quality images showing real-world human faces from different angles at a resolution of $1024\times 1024$.
To obtain the pose of a face we use an off-the-shelf pose estimator~\cite{Deng2019Accurate3F}.

\noindent\textbf{\afhq.} The \afhq-Cats dataset~\cite{choi2020starganv2,KarrasNeurIPS2021} consists of  5,065 images showing faces of cats from different views at a resolution of $512\times 512$. We augment the dataset by horizontal flips and obtain the pose of a cat's face via an off-the-shelf cat face landmark predictor~\cite{CatHipsterizer} and OpenCV's perspective-n-point algorithm.

\noindent\textbf{MetFaces.} The MetFaces dataset~\cite{Karras2020TrainingGA} consists of 1,336 images showing high-quality faces extracted from the collection of the Metropolitan Museum of Art.\footnote{\url{https://metmuseum.github.io/}} To augment the dataset we again use horizontal flips. To obtain the pose of a face we use an off-the-shelf pose estimator~\cite{Deng2019Accurate3F}.

\subsection{Evaluation Metrics}\label{sec: eval metrics}
We follow prior work and assess the obtained results using common metrics:

\noindent\textbf{2D GAN metrics.} We report Fr\'echet Inception Distance (FID)~\cite{Heusel2017GANsTB} and Kernel Inception Distance (KID)~\cite{Binkowski2018DemystifyingMG}, computed by using 50k artificially generated images that were rendered from random poses and 1)  50k real images for FFHQ; 2) all 10,130 real images in the $x$-flip-augmented dataset for \afhq-Cat~\cite{choi2020starganv2}. 

\noindent\textbf{Identity similarity (ID).} Following EG3D~\cite{ChanARXIV2021}, we also evaluate the level of facial identity preservation. Concretely, we first generate 1024 MPI-like representations ${\cal M}$. For each representation  we then compute the identity cosine similarity between two  views rendered from random poses using ArcFace~\cite{Deng2019ArcFaceAA,serengil2020lightface}. 

\noindent\textbf{Depth accuracy (Depth).} Similar to~\cite{ChanARXIV2021,Shi2021Lifting2S}, we also assess geometry and depth accuracy. For this we utilize a pre-trained face reconstruction model~\cite{Deng2019Accurate3F} to provide facial area mask and pseudo ground-truth depth map $\widehat{D}_{v_\texttt{tgt}}$. We report the MSE error between our rendered depth $D_{v_\texttt{tgt}}$ (\equref{eq: mpi over depth}) and $\widehat{D}_{v_\texttt{tgt}}$ on areas constrained by the face mask. Note, following prior work, we normalize both depth maps to zero-mean and unit-variance. The  result is obtained by averaging over 1024 representations ${\cal M}$.

\noindent\textbf{Pose accuracy (Pose).} Following~\cite{ChanARXIV2021,Shi2021Lifting2S}, we also study the 3D geometry's pose accuracy. Specifically, for each MPI, we utilize a pose predictor~\cite{Deng2019Accurate3F} to estimate the yaw, pitch, and roll of a rendered image. The predicted pose is then compared to the pose used for rendering via the MSE. The reported result is  averaged over 1024 representations ${\cal M}$. 
Notably but in hindsight expected, 2D metrics, as well as ID, lack the ability to capture 3D errors which we will show next.

\begin{table}[!t]
\renewcommand{\arraystretch}{1.0}
\begin{adjustwidth}{0.0cm}{}
\renewcommand\theadfont{}
\centering
\setlength\aboverulesep{0pt}
\setlength\belowrulesep{0pt}
\captionsetup{width=\linewidth}
\caption{
\textbf{Speed comparison.}
`--' indicates that corresponding papers do not report this result.
\textbf{1) Training:}
EG3D, GRAM, StyleNeRF and GMPI report training time when using 8 Tesla V100 GPUs.
pi-GAN uses two RTX 6000 GPUs or a single RTX 8000 GPU.
For GMPI, the results reported in this paper come from 3/5/11-hour training for a resolution of $256^2$/$512^2$/$1024^2$ with initialization from official pretrained-checkpoints.
\textbf{2) Inference:}
we measure frames-per-second (FPS) for each model. GMPI uses 96 planes.
GRAM reports speed by utilizing a specified mesh rasterizer~\cite{Laine2020ModularPF} while the others use pure forward passes.
EG3D uses an RTX 3090 GPU.
GRAM and StyleNeRF use a Tesla V100 GPU, which we also utilize to run GIRAFFE, pi-GAN, LiftedGAN, and our GMPI.
We observe: GMPI is quick to train and renders the fastest among all approaches.
}
\label{tab: speed}
\begin{adjustbox}{width=\columnwidth,center}
{
\begin{threeparttable}
\begin{tabular}{lll|rrrrrr|r} 
\toprule
 & Res. & Unit & {\scriptsize GIRAFFE~\cite{NiemeyerCVPR2021}} & {\scriptsize pi-GAN~\cite{ChanCVPR20201}} & {\scriptsize LiftedGAN~\cite{Shi2021Lifting2S}} & {\scriptsize EG3D$^\dagger$~\cite{ChanARXIV2021}} & {\scriptsize GRAM$^\dagger$~\cite{Deng2021GRAMGR}} & {\scriptsize StyleNeRF$^\dagger$~\cite{GuARXIV2021}} & {\scriptsize GMPI} \\
\midrule
{\scriptsize Train} & - & Time$\downarrow$ & --    & 56h  &  --  & 8.5d & 3-7d & 3d & \textbf{3/5/11h}\\
\midrule
\addlinespace[1pt]
\parbox[t]{2mm}{\multirow{3}{*}{\rotatebox[origin=c]{90}{\scriptsize\makecell{Inference}}}}
& $256^2$  & FPS$\uparrow$ & 250 & 1.63  & 25 & 36   & 180 & 16 & \textbf{328}$^\ast$ \\
& $512^2$  & FPS$\uparrow$ & --  & 0.41  & -- & 35   & --  & 14 & \textbf{83.5}$^\ast$ \\
& $1024^2$ & FPS$\uparrow$ & -- & 0.10 & -- & -- & --  & 11 & \textbf{19.4}$^\ast$ \\
\toprule
\end{tabular}
\begin{tablenotes}
\item[$\dagger$] We quote results from their papers.
\item[$\ast$] We report the  rendering speed. GMPI, different from StyleNeRF, pi-GAN, and LiftedGAN, only needs a single forward pass to generate the scene representation. Further rendering doesn't involve the generator. We hence follow EG3D to report  inference FPS without forward pass time that is 82.34 ms, 99.97 ms, and 115.10 ms for 96 planes at $256^2$, $512^2$, and $1024^2$ on a V100 GPU.
\end{tablenotes}
\end{threeparttable}
}
\end{adjustbox}
\end{adjustwidth}
\end{table}

\begin{table}[!t]
\renewcommand{\arraystretch}{1.0}
\captionsetup{width=\linewidth}
\caption{
\textbf{Quantitative results.}
GMPI uses 96 planes and no truncation trick~\cite{Karras2019ASG,Brock2019LargeSG} is applied.
`--' indicates that the corresponding paper does not report this result.
KID is reported in KID$\times100$.
For GIRAFFE, pi-GAN, and LiftedGAN, we quote results from~\cite{ChanARXIV2021}.
Compared to baselines, GMPI achieves on-par or even better performance despite much faster training as reported in~\tabref{tab: speed}. Moreover, GMPI can produce high-resolution images which most baselines fail to achieve, demonstrating the flexibility.
}
\label{tab: qunatitative}
\begin{adjustwidth}{0.0cm}{}
\renewcommand\theadfont{}
\centering
\setlength\aboverulesep{0pt}
\setlength\belowrulesep{0pt}
\setlength{\tabcolsep}{3pt}
{
\small
\begin{threeparttable}
\begin{tabular}{lll@{\hskip 0.3em}r@{\hskip 0.65em}r@{\hskip 0.65em}r@{\hskip 0.65em}r@{\hskip 0.65em}r@{\hskip 0.65em}r@{\hskip 0.65em}r} 
\toprule
& & & \multicolumn{5}{c}{FFHQ} & \multicolumn{2}{c}{\afhq-Cat} \\
\cmidrule(r){4-8} \cmidrule(){9-10}
 &  & & FID$\downarrow$ & KID$\downarrow$  & ID$\uparrow$ & Depth$\downarrow$ & Pose$\downarrow$ & FID$\downarrow$ & KID$\downarrow$ \\
\midrule
\parbox[t]{2mm}{\multirow{9}{*}{\rotatebox[origin=c]{90}{\scriptsize\makecell{$256^2$}}}}
& 1 & GIRAFFE~\cite{NiemeyerCVPR2021}                 & 31.5 & 1.992 & 0.64 & 0.94 & 0.089 & 16.1 & 2.723 \\
& 2 & pi-GAN $128^2$~\cite{ChanCVPR20201}             & 29.9 & 3.573 & 0.67 & 0.44 & 0.021 & 16.0 & 1.492 \\
& 3 & LiftedGAN~\cite{Shi2021Lifting2S}               & 29.8 & --    & 0.58 & 0.40 & 0.023 & -- & -- \\
& 4 & GRAM$^\dagger$~\cite{Deng2021GRAMGR}            & 29.8 & 1.160 & --   & --   & --   & -- & -- \\
& 5 & StyleSDF$^\dagger$~\cite{OrEl2021StyleSDFH3}    & 11.5 & 0.265 & --   & -- & --   & 12.8$^\ast$ & 0.447$^\ast$ \\
& 6 & StyleNeRF$^\dagger$~\cite{GuARXIV2021}          & 8.00 & 0.370 & --   & --   & --   & 14.0$^\ast$ & 0.350$^\ast$ \\
& 7 & CIPS-3D$^\dagger$~\cite{ZhouARXIV2021}                    & 6.97 & 0.287 & --   & --   & --   & -- & -- \\
& 8 & EG3D$^\dagger$~\cite{ChanARXIV2021}             & 4.80 & 0.149 & 0.76 & 0.31 & 0.005 & 3.88 & 0.091 \\
& 9 & \textbf{GMPI}  &  11.4  & 0.738 & 0.70  & 0.53 & 0.004 & n/a$^\mathsection$ & n/a$^\mathsection$ \\
\midrule
\addlinespace[1pt]
\parbox[t]{2mm}{\multirow{3}{*}{\rotatebox[origin=c]{90}{\scriptsize\makecell{$512^2$}}}}
& 10 & EG3D$^\dagger$~\cite{ChanARXIV2021}   & 4.70 & 0.132  & 0.77 & 0.39 & 0.005 & 2.77 & 0.041 \\
& 11 & StyleNeRF$^\dagger$~\cite{GuARXIV2021} & 7.80 & 0.220 & -- & -- & -- & 13.2$^\ast$ & 0.360$^\ast$ \\
& 12 & \textbf{GMPI}  & 8.29 & 0.454 & 0.74 & 0.46 & 0.006 & 7.79 & 0.474   \\
\midrule
\addlinespace[1pt]
\parbox[t]{2mm}{\multirow{3}{*}{\rotatebox[origin=c]{90}{\scriptsize\makecell{$1024^2$}}}}
& 13 & CIPS-3D$^\dagger$~\cite{ZhouARXIV2021}          & 12.3 & 0.774  & --  & -- & -- & -- & -- \\
& 14 & StyleNeRF$^\dagger$~\cite{GuARXIV2021} & 8.10 & 0.240  & --  & -- & -- & -- & -- \\
& 15 & \textbf{GMPI} & 7.50 & 0.407  & 0.75 & 0.54 & 0.007 & n/a$^\mathsection$ & n/a$^\mathsection$ \\

\toprule
\end{tabular}
\begin{tablenotes}\scriptsize
\item[$\dagger$] We quote results from their papers.
\item[$\ast$] Performance is reported on the whole \afhq~dataset instead of only cats.
\item[$\mathsection$] No GMPI results on \afhq~for Row 9 and 13 since there are no available pre-trained \stylegan~checkpoints for \afhq~with a resolution of $256^2$ or $1024^2$.
\end{tablenotes}
\end{threeparttable}
}
\end{adjustwidth}
\end{table}

\begin{table}[t]
\renewcommand{\arraystretch}{1.0}
\begin{adjustwidth}{0.0cm}{}
\captionsetup{width=\linewidth}
\caption{
\textbf{Ablation studies.}
We evaluate at a resolution of $512^2$.
\texttt{DPC} refers to discriminator pose conditioning (\secref{sec:dpc}), 
$\mathcal{F}_{\alpha_i}^h$ refers to the plane-specific feature introduced in~\equref{eq: F_alpha}, and 
\texttt{Shading} indicates the shading-guided training discussed in~\secref{sec:ma}.
\texttt{\#planes} denotes the number of planes we used during evaluation. Note, rows (b) and (c) can only use 32 planes during training and inference. Therefore, to make the comparison fair, all ablations use 32 planes. We provide an additional 96-plane result for our full model, which is used in~\tabref{tab: qunatitative}.
Please check~\secref{sec: ablation} and~\figref{fig: ablation} for a detailed discussion regarding this table as we find that 2D metrics,~\eg,~ FID/KID as well as ID can be misleading when evaluating 3D generative models.
}
\label{tab: ablations}
\begin{adjustbox}{width=0.9\columnwidth,center}
\renewcommand\theadfont{}
\centering
\setlength\aboverulesep{0pt}
\setlength\belowrulesep{0pt}
\setlength{\tabcolsep}{3pt}
{
\small
\begin{tabular}{lccccc@{\hskip 0.3em}r@{\hskip 0.65em}r@{\hskip 0.65em}r@{\hskip 0.65em}r@{\hskip 0.65em}r@{\hskip 0.8em}l@{\hskip 0.65em}l} 
\toprule
& \multirow{2}{*}{\texttt{\#planes}} & \multirow{2}{*}{\texttt{DPC}} & \multirow{2}{*}{$\mathcal{F}_{\alpha_i}^h$} & \multirow{2}{*}{\texttt{Shading}} & \multicolumn{5}{c}{FFHQ} & \multicolumn{2}{c}{\afhq-Cat} \\
\cmidrule(r){6-10} \cmidrule(){11-12}
 & & & & & FID$\downarrow$ & KID$\downarrow$  & ID$\uparrow$ & Depth$\downarrow$ & Pose$\downarrow$ & FID$\downarrow$ & KID$\downarrow$ \\
\midrule
(a)   & 32 &         &        & & 6.64 & 0.368  & \textbf{0.91} & 2.043 & 0.060 & 4.29 & 0.199 \\ 
\midrule
(b)   & 32 &         &        & & 4.61 & 0.167 & 0.89 & 2.190 & 0.062 & 4.00 & 0.166 \\
(c)   & 32 & \cmark  &        & & 7.98 & 0.347 & 0.75 & 0.501 & \textbf{0.006} & 7.13 & 0.385 \\
(d)   & 32 &         & \cmark & & \textbf{4.35} & \textbf{0.150} & 0.89 & 2.140 & 0.061 & 3.70 & 0.132 \\

\midrule
(e) & 32 & \cmark & \cmark &  & 7.19 & 0.313 & 0.73 & 0.462 & \textbf{0.006} & 7.54 & 0.433  \\

\midrule

(f) & 32 & \cmark & \cmark & \cmark & 7.40 & 0.337 & 0.74 & \textbf{0.457} & \textbf{0.006} & 7.93 & 0.489 \\
(g) & 96 & \cmark & \cmark & \cmark & 8.29 & 0.454 & 0.74 & \textbf{0.457} & \textbf{0.006} & 7.79 & 0.474 \\

\toprule
\end{tabular}
}
\end{adjustbox}
\end{adjustwidth}
\end{table}

\subsection{Results}
\label{sec:results}
We provide speed comparison in~\tabref{tab: speed} and a quantitative evaluation in~\tabref{tab: qunatitative}.
With faster training, GMPI achieves on-par or better performance than start-of-the-art when evaluating on $256^2$ images and can generate high-resolution results up to $1024^2$ which most baselines fail to produce.
Specifically, GMPI results on resolutions  $256^2$, $512^2$, and $1024^2$ are reported after 3/5/11-hours of training.
Note, the pretrained \stylegan~initialization for FFHQ (see~\secref{sec: training}) requires training for 1d 11h ($256^2$), 2d 22h ($512^2$), and 6d 03h ($1024^2$)  with 8 Tesla V100 GPUs respectively, as reported in the official repo.\cref{footnote: stylegan repo}
In contrast, EG3D, GRAM, and StyleNeRF require training of at least three days. 
At a resolution of $256^2$,
1) GMPI outperforms GIRAFFE, pi-GAN, LiftedGAN, and GRAM on FID/KID while outperforming StyleSDF on FID;
2) GMPI demonstrates better identity similarity (ID) than GIRAFFE, pi-GAN, and LiftedGAN; 
3) GMPI outperforms GIRAFFE regarding depth;
4) GMPI performs best among all baselines on pose accuracy.
Overall, GMPI demonstrates that it is a flexible architecture which achieves 3D-awareness with an affordable training time.

\begin{figure}[!t]
    \centering
    \captionsetup[subfigure]{aboveskip=1pt}
        \centering
        \includegraphics[width=0.9\textwidth]{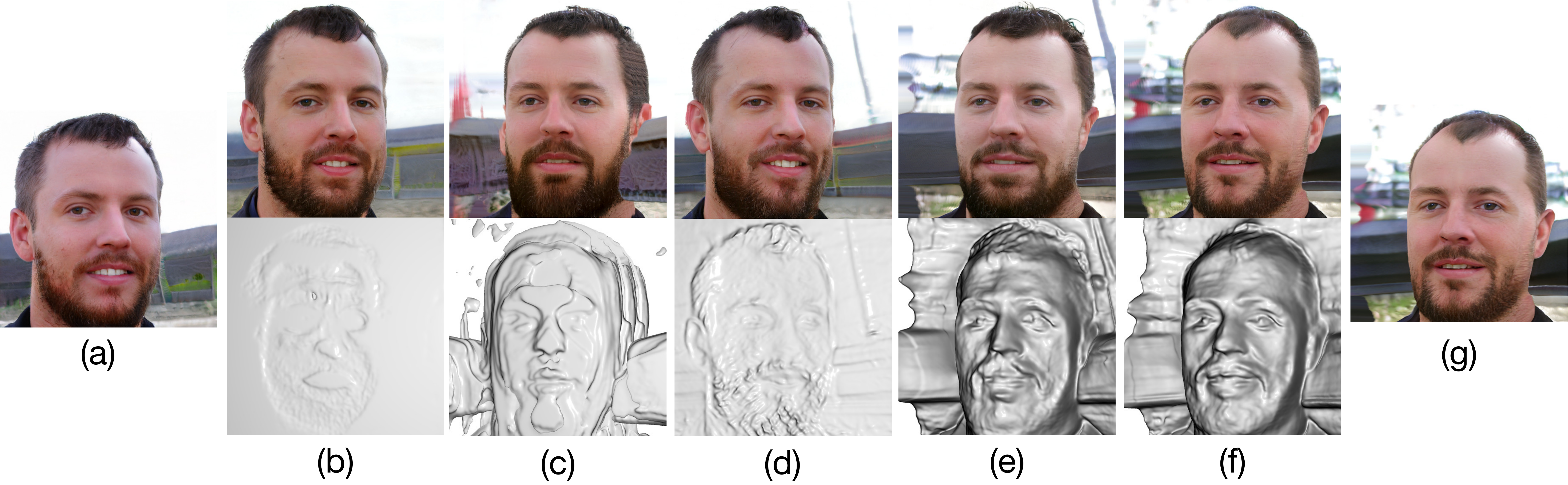}
        \captionsetup{width=\textwidth}
        \caption{
        The images in this figure correspond the the ablation studies in \tabref{tab: ablations}. Panels (a)-(g) correspond to \tabref{tab: ablations}'s rows (a)-(g). All results are generated from the same latent code $\bm{z}$.
        The face is rendered with a camera positioned to the right of the subject, \ie, the rendered face should look to the left of the viewer. 
        (a) is the 2D image produced by the pre-trained \stylegan{}.
        Note how GMPI becomes 3D-aware in (e), and generates geometry and texture that is occluded in the pre-trained \stylegan~image. 
        (e)~\vs~(f): shading-guided training (\secref{sec:ma}) alleviates geometric artifacts such as concavities in the forehead.
        (f)~\vs~(g): the ability to use more planes during inference (\secref{sec:mib}) reduces ``stair step'' artifacts, visible on the cheek and the ear.
        }
        \label{fig: ablation compare}
    \label{fig: ablation}
\end{figure}

\begin{figure}[!t]
    \centering
    \captionsetup[subfigure]{aboveskip=1pt}
    \begin{subfigure}{0.9\textwidth}
        \centering
        \includegraphics[width=\textwidth]{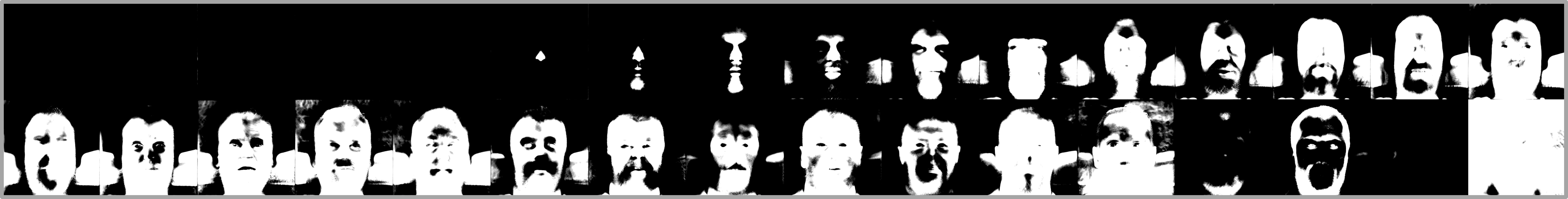}
        \captionsetup{width=\textwidth}
        \caption{
        Generated alpha maps for~\tabref{tab: ablations} row c. The network fails to produce realistic structures. 
        }
        \label{fig: ablation alpha no F}
    \end{subfigure}%
    \hfill
    \begin{subfigure}{0.9\textwidth}
        \centering
        \includegraphics[width=\textwidth]{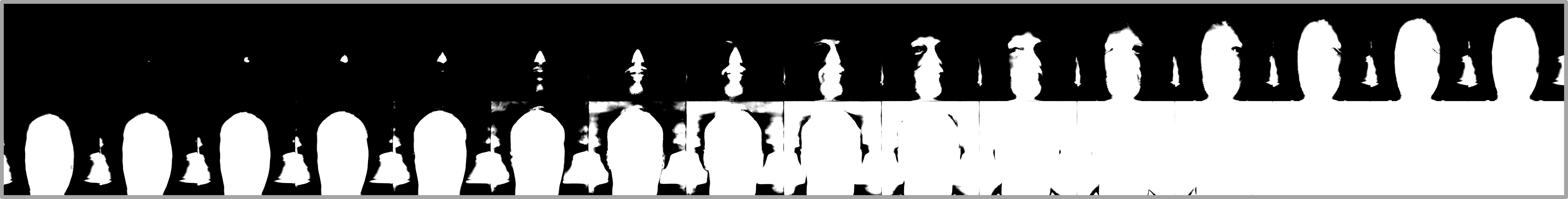}
        \captionsetup{width=\textwidth}
        \caption{
        Generated alpha maps for~\tabref{tab: ablations} row g. Note how the alpha maps are more structured and more closely correspond to a human’s face structure. 
        }
        \label{fig: ablation alpha ours}
    \end{subfigure}%
    \caption{
    \textbf{Qualitative results accompanying~\tabref{tab: ablations}.}
    For alpha maps, from top to bottom, left to right, we show $\alpha_1$ to $\alpha_L$ respectively (\equref{eq: mpi define}): the whiter, the denser the occupancy. Grey boundaries are added for illustration.
    }
    \label{fig: ablation_alpha}
\end{figure}

\subsection{Ablation Studies}\label{sec: ablation}

In order to show the effects of various design choices, we run ablation studies and selectively drop out the discriminator conditioned on pose (\texttt{DPC}),  the plane-specific feature $\mathcal{F}_{\alpha_i}^h$, and shading.  The following  quantitative (\tabref{tab: ablations}) and qualitative (\figref{fig: ablation} and \figref{fig: ablation_alpha}) studies were run using a resolution of $512^2$ to set baseline metrics, and answer the following questions:

\noindent\textbf{Baseline condition} {(\tabref{tab: ablations} row a)} --  a pre-trained \stylegan{} texture without transparency ($\alpha_i \forall i$ is set to 1).  This representation is not truly 3D, as it is rendered as a textured block in space.  It is important to note that the commonly used metrics for 2D GAN evaluation, \ie, FID, KID, as well as ID yield good results. They are hence not sensitive to 3D structure. In contrast, the depth and pose metrics do capture the lack of 3D structure in the scene.

\noindent \textbf{Can a na\"ive MPI generator learn 3D?} {(\tabref{tab: ablations} row b)} -- a generator which only uses $\mathcal{F}^h$  (\equref{eq: stylegan C^h}) without being trained with pose-conditioned discriminator, plane-specific features, or a shading loss.   
The Depth (2.190~\vs~2.043) and Pose (0.062~\vs~0.060) metrics show that this design fails, yielding results similar to the baseline (row a).

\noindent \textbf{Does \texttt{DPC} alone enable 3D-awareness?} {(\tabref{tab: ablations} row c)} -- we now use discriminator pose conditioning (\texttt{DPC}) while ignoring plane-specific features $\mathcal{F}_{\alpha_i}^h$. 
This design improves  the Depth (0.501~\vs~2.190) and Pose (0.006~\vs~0.062) metrics. However upon closer inspection of the alpha maps shown in \figref{fig: ablation alpha no F} we observe that the generator fails to produce realistic structures. The reason that this design performs well on Depth and Pose metrics is primarily due to these two metrics only evaluating the surface while not considering the whole volume.

\noindent \textbf{Does $\mathcal{F}_{\alpha_i}^h$ alone enable 3D-awareness?} {(\tabref{tab: ablations} row d)} --  we use only the plane-specific feature $\mathcal{F}_{\alpha_i}^h$ without \texttt{DPC}. Geometry related metrics (Depth and Pose) perform poorly, which is corroborated by \figref{fig: ablation compare} (d), indicating an inability to model 3D information. 

\noindent \textbf{Do \texttt{DPC} and $\mathcal{F}_{\alpha_i}^h$ enable 3D-awareness?} {(\tabref{tab: ablations} row e)} -- 
we combine both plane-specific features $\mathcal{F}_{\alpha_i}^h$ and \texttt{DPC}. This produce good values for Depth and Pose metrics. Visualizations in \figref{fig: ablation compare}e, and~\figref{fig: ablation alpha ours} verify that GMPI successfully generates 3D-aware content. However, the FID/KID, as well as ID values are generally worse than non-3D-aware generators (row a-e). 

\noindent \textbf{Does shading improve 3D-awareness?} {(\tabref{tab: ablations} row f)} -- we use plane-specific features $\mathcal{F}_{\alpha_i}^h$, \texttt{DPC} and shading loss.
As discussed in~\secref{sec: overview}, \texttt{DPC} and $\mathcal{F}_{\alpha_i}^h$ are sufficient to make a 2D GAN 3D-aware. However, inspecting the geometry reveals artifacts such as the concave forehead in~\figref{fig: ablation compare}e. Shading-guided rendering tends to alleviate these issues (\figref{fig: ablation compare}e~\vs~f) while not harming the quantitative results (\tabref{tab: ablations}'s row e~\vs~f).

\noindent \textbf{Can we reduce aliasing artifacts?} {(\tabref{tab: ablations} row g)} -- we use  plane-specific features $\mathcal{F}_{\alpha_i}^h$, \texttt{DPC}, shading loss and 96 planes. 
Due to the formulation of $\mathcal{F}_{\alpha_i}^h$, we can generate an arbitrary number of planes during inference, which helps avoid ``stair step'' artifacts that can be observed in ~\figref{fig: ablation compare}f~\vs~g. %

\noindent\textbf{Qualitative results} are shown in \cref{fig:teaser}. Please see the appendix for more.

\section{Conclusion}
\label{sec:conc}
To identify what is really needed to make a 2D GAN 3D-aware we develop generative multiplane images (GMPIs) which \emph{guarantee view-consistency}. GMPIs show that a \stylegan~can be made 3D-aware by  1) adding a multiplane image style  branch which generates a set of alpha maps conditioned on their depth in addition to a single image, both of which are used for rendering via an end-to-end differentiable warping and alpha compositing; and by 2) ensuring that the discriminator  is conditioned on the pose. 
We also identify shortcomings of classical evaluation metrics used for 2D image generation. 
We hope that the simplicity of GMPIs inspires future work to fix  limitations such as occlusion reasoning.

$\newline$
\noindent\textbf{Acknowledgements:} We thank Eric Ryan Chan for discussion and providing processed \afhq-Cats dataset. Supported in part by NSF grants 1718221, 2008387, 2045586, 2106825, MRI \#1725729, NIFA award 2020-67021-32799.

\clearpage
\bibliographystyle{splncs04}
\bibliography{egbib}

\clearpage
\beginsupplement
\appendix

\section*{\Large\centering Supplementary Material: \\Generative Multiplane Images:\\Making a 2D GAN 3D-Aware}

\renewcommand{\thesection}{\Alph{section}}

This supplementary is structured as follows:
\begin{itemize}
    \item \secref{supp sec: gmpi render}: Details of the differentiable rendering in GMPI;
    \item \secref{supp sec: quant}: Additional quantitative results;
    \item \secref{supp sec: implement}: Implementation details;
    \item \secref{supp sec: qualitative}: More qualitative results.
\end{itemize}

\section{Differentiable Rendering in GMPI}\label{supp sec: gmpi render}

In~\secref{sec:r}, we obtain the desired image $I_{v_\texttt{tgt}}$ which illustrates the generated MPI representation 
${\cal M} = \left\{C, \{\alpha_1, \dots, \alpha_L\}\right\}$ 
from the user-specified target view $v_\texttt{tgt}$ in two steps: 1) a warping step transforms the  representation $\mathcal{M}$ from its canonical pose $v_\texttt{cano}$ to the target pose $v_\texttt{tgt}$; 2) a compositing step combines the planes into the desired image $I_{v_\texttt{tgt}}$. Importantly, both steps entail easy computations which are end-to-end differentiable such that they can be included into any generator.
Here we provide details.

\noindent\textbf{Warping.} We warp the  RGB image and the alpha map of the $i^\text{th}$ plane from the canonical view to the target view via
\begin{align}
    (C_i^\prime, \alpha_i^\prime) = \mathcal{H}_{i, v_\texttt{cano} \rightarrow v_\texttt{tgt}}(C, \alpha_i, d_i).
    \label{eq:transform}
\end{align}
Here, $\mathcal{H}_{i, v_\texttt{cano} \rightarrow v_\texttt{tgt}}$ represents the homography operation. Essentially, the homography $\mathcal{H}_{i, v_\texttt{cano} \rightarrow v_\texttt{tgt}}$ specifies a mapping: for each pixel coordinate $(p_x^\prime, p_y^\prime)$ in the image $C_i^\prime$ and in the alpha map $\alpha_i^\prime$ of the target view $v_\texttt{tgt}$,  we obtain corresponding coordinates $(p_x, p_y)$ in the image $C$  and in the alpha map $\alpha_i$ of the canonical view $v_\texttt{cano}$. Bilinear sampling is applied on $(p_x, p_y)$ to obtain values for the pixel locations $(p_x^\prime, p_y^\prime)$. Concretely,
\begin{align}
    \begin{bmatrix}
    p_x& p_y& 1
    \end{bmatrix}^\top = K_{v_\texttt{cano}} \left( R_{v_\texttt{tgt} \rightarrow v_\texttt{cano}} - \frac{\bm{t}_{v_\texttt{tgt} \rightarrow v_\texttt{cano}} \bm{n}^\top}{b_i} \right) K_{v_\texttt{tgt}}^{-1} \begin{bmatrix}
    p_x^\prime& p_y^\prime& 1
    \end{bmatrix}^\top, \label{eq: mpi warp}
\end{align}
where $\bm{n} \in \mathbb{R}^3$ is the normal of the plane defined in target camera coordinate system $v_\texttt{tgt}$, which is identical for all planes. $b_i$ is the depth of the plane from the target camera $v_\texttt{tgt}$. We let $K_{v_\texttt{cano}} \in \mathbb{R}^{3\times3}$ and $K_{v_\texttt{tgt}} \in \mathbb{R}^{3\times3}$ refer to the intrinsic matrices of the canonical view $v_\texttt{cano}$ and the target view $v_\texttt{tgt}$. Further, $R_{v_\texttt{tgt} \rightarrow v_\texttt{cano}} \in \mathbb{R}^{3\times3}$ and $\bm{t}_{v_\texttt{tgt} \rightarrow v_\texttt{cano}} \in \mathbb{R}^{3\times1}$ are the rotation  and the translation  from $v_\texttt{tgt}$ to $v_\texttt{cano}$.

\noindent\textbf{Alpha Compositing.} Given the warped image $C_i^\prime$ and the warped alpha map $\alpha_i^\prime$ for each plane $i$, we compute the final rendered 2D image $I_{v_\texttt{tgt}}$ via
\begin{align}
    I_{v_\texttt{tgt}} = \sum\limits_{i=1}^L \left( C_i^\prime \cdot \alpha_i^\prime \cdot \prod\limits_{j=1}^{i-1} (1 - \alpha_j^\prime) \right). \label{eq: mpi over}
\end{align}

Similarly, we approximate depth $D_{v_\texttt{tgt}}$ via
\begin{align}
    D_{v_\texttt{tgt}} = \sum\limits_{i=1}^L \left( b_i \cdot \alpha_i^\prime \cdot \prod\limits_{j=1}^{i-1} (1 - \alpha_j^\prime) \right), \label{eq: mpi over depth}
\end{align}
where $b_i$ is the distance mentioned in~\equref{eq: mpi warp}.
Notably, the combination of \equref{eq: mpi define}, \equref{eq:transform} and \equref{eq: mpi over} is end-to-end differentiable and hence straightforward to integrate into a generator. Importantly, the computations are also extremely efficient as only simple matrix multiplications are involved. It is hence easy to augment an existing generator like the one in \stylegan.

\section{Additional Quantitative Results}\label{supp sec: quant}

\begin{table}[!t]
\renewcommand{\arraystretch}{}
\begin{adjustwidth}{0.0cm}{}
\renewcommand\theadfont{}
\centering
\setlength\aboverulesep{0pt}
\setlength\belowrulesep{0pt}
{
\captionsetup{width=\linewidth}
\caption{
\textbf{Comparing representations on FFHQ.}
$1^\text{st}$ and $3^\text{rd}$ rows are copied from~\tabref{tab: qunatitative}'s $3^\text{rd}$ and $9^\text{th}$ rows in the main text. Row 2 and 3 infer with 96 planes.
Row 1: depth to renderer; Row 2: depth to MPI to renderer; Row 3: MPI to renderer.
}
\label{supp tab: representation}
\begin{threeparttable}
\begin{tabular}{lll@{\hskip 0.3em}r@{\hskip 0.65em}r@{\hskip 0.65em}r@{\hskip 0.65em}r@{\hskip 0.65em}r@{\hskip 0.65em}r@{\hskip 0.65em}r} 
\toprule
 & Row & & FID$\downarrow$ & KID{\scriptsize $\times100\downarrow$}  & ID$\uparrow$ & Depth$\downarrow$ & Pose$\downarrow$ \\
\midrule
\parbox[t]{2mm}{\multirow{6}{*}{\rotatebox[origin=c]{90}{\scriptsize\makecell{$256^2$}}}}
& 1 & LiftedGAN [58]              & 29.8 & --    & 0.58 & \textbf{0.40} & 0.023  \\

\cmidrule(r){2-8}
& 2-1 & \texttt{D2A} {\scriptsize($\epsilon=1/64$)}  & 13.4 & 0.920 & 0.69 & 0.60 & \textbf{0.004} \\
& 2-2 & \texttt{D2A} {\scriptsize($\epsilon=1/128$)} & 13.5 & 0.867 & \textbf{0.70} & 0.60 & \textbf{0.004} \\
& 2-3 & \texttt{D2A} {\scriptsize($\epsilon=1/256$)}  & 11.7 & \textbf{0.644} & \textbf{0.70} & 0.63 & 0.005 \\
& 2-4 & \texttt{D2A} {\scriptsize($\epsilon=1/512$)}  & 12.6 & 0.684 & 0.69 & 0.62 & 0.005 \\

\cmidrule(r){2-8}
& 3 & \textbf{GMPI}  &  \textbf{11.4}  & 0.738 & \textbf{0.70}  & 0.53 & \textbf{0.004} \\
\toprule
\end{tabular}
\end{threeparttable}
}
\end{adjustwidth}
\end{table}

\subsection{Comparisons of Representations}\label{supp sec: repre}

We provide additional ablations of the plane representation. Specifically, we consider the following three ablations:

\noindent\textbf{Depth map to renderer:} We let the generator predict a single channel depth map (instead of multiple-channel alpha maps) and use a differentiable renderer to supervise the transformed image. LiftedGAN~\cite{Shi2021Lifting2S} serves as this ablation.

\noindent\textbf{Depth maps to alpha maps to renderer:} We compute multiple alpha maps from a predicted single-channel depth map in two steps:
1) we predict a depth map $\texttt{Depth} \in \mathbb{R}^{H \times W}$ in normalized coordinates; 2) we generate $L$ alpha maps from \texttt{Depth} by computing the alpha value for a  pixel $[x, y]$ on the $i$-th alpha map $\alpha_i$ via:
\begin{align}
    \alpha_i [x, y]=\min \left( 1, \max \left(0, \frac{ d_i - (\texttt{Depth}[x, y] - \epsilon)}{ 2 \epsilon} \right) \right),\nonumber
\end{align}
where $d_i$ is $\alpha_i$'s depth. Essentially,  alpha values linearly increase from 0 to 1 within the range  $[\texttt{Depth}[x, y] - \epsilon, \texttt{Depth}[x, y] + \epsilon]$. Any depth closer than $\texttt{Depth}[x, y] - \epsilon$ is set to 0. Any depth further than $\texttt{Depth}[x, y] + \epsilon$ is set to 1. We call this representation \texttt{D2A}.

\noindent\textbf{Alpha maps to renderer:} We directly predict multiple-channel alpha maps. Our GMPI serves as this ablation.

In~\tabref{supp tab: representation} we report the results. For \texttt{D2A}, we ablate several values of $\epsilon$ and find that GMPI always outperforms on FID and depth metrics, verifying the fidelity of th texture and the  high-quality of the geometry.

\subsection{Depth Score Analysis}\label{supp sec: depth}

\begin{table}[t]
\renewcommand{\arraystretch}{1.0}
\begin{adjustwidth}{0.0cm}{}
\captionsetup{width=\linewidth}
\caption{
\textbf{Ablation studies on truncation level $\psi$ during inference on FFHQ.}
All results use the same checkpoint trained with \texttt{DPC}, plane-specific features $\mathcal{F}_{\alpha_i}^h$, and shading-guided training.
We evaluate with 96 planes.
We show more digits for `Depth' and `Pose' than we use in~\tabref{tab: qunatitative} to emphasize differences.
With lower $\psi$, GMPI produces geometries with fewer  artifacts (Depth) at the cost of less variety (FID/KID).
}
\label{supp tab: depth score}
\renewcommand\theadfont{}
\centering
\setlength\aboverulesep{0pt}
\setlength\belowrulesep{0pt}
\begin{adjustbox}{width=\columnwidth,center}
\setlength{\tabcolsep}{3pt}
{
\small
\begin{tabular}{lc@{\hskip 0.3em}r@{\hskip 0.65em}r@{\hskip 0.65em}r@{\hskip 0.65em}r@{\hskip 0.65em}r @{\hskip 0.65em}r@{\hskip 0.65em}r@{\hskip 0.65em}r@{\hskip 0.65em}r@{\hskip 0.65em}r @{\hskip 0.65em}r@{\hskip 0.65em}r@{\hskip 0.65em}r@{\hskip 0.65em}r@{\hskip 0.65em}r} 
\toprule
& \multirow{2}{*}{$\psi$} & \multicolumn{5}{c}{$256^2$} & \multicolumn{5}{c}{$512^2$} & \multicolumn{5}{c}{$1024^2$} \\
\cmidrule(r){3-7} \cmidrule(r){8-12} \cmidrule(r){13-17} 
 & & FID$\downarrow$ & KID$\downarrow$  & ID$\uparrow$ & Depth$\downarrow$ & Pose$\downarrow$ & FID$\downarrow$ & KID$\downarrow$  & ID$\uparrow$ & Depth$\downarrow$ & Pose$\downarrow$ & FID$\downarrow$ & KID$\downarrow$  & ID$\uparrow$ & Depth$\downarrow$ & Pose$\downarrow$ \\
\midrule
(a) & 1.0 & \textbf{11.4}  & \textbf{0.738} & \textbf{0.70}  & 0.533 & 0.0042 & \textbf{8.29} & \textbf{0.454} & \textbf{0.74} & 0.455 & 0.0056 & \textbf{7.50} & \textbf{0.407} & \textbf{0.75} & 0.535 & 0.0068 \\
(b) & 0.9 & 12.6 & 0.837 & 0.70 & 0.515 & 0.0037 & 10.1 & 0.588 & 0.74 & 0.421 & 0.0047 & 8.83 & 0.496 & 0.75 & 0.490 & 0.0058 \\
(c) & 0.8 & 15.4 & 1.079 & 0.70 & 0.497 & 0.0033 & 13.7 & 0.885 & 0.73 & 0.386 & 0.0039 & 11.5 & 0.691 & 0.74 & 0.449 & 0.0049 \\
(d) & 0.7 & 20.3 & 1.510 & 0.70 & 0.477 & 0.0029 & 19.8 & 1.391 & 0.73 & 0.354 & 0.0032 & 16.0 & 1.024 & 0.73 & 0.408 & 0.0040 \\
(e) & 0.6 & 27.8 & 2.172 & 0.70 & 0.459 & 0.0026 & 28.8 & 2.160 & 0.73 & 0.326 & 0.0025 & 22.8 & 1.510 & 0.73 & 0.368 & 0.0031 \\
(f) & 0.5 & 38.5 & 3.112 & 0.70 & \textbf{0.445} & \textbf{0.0023} & 41.1 & 3.249 & 0.72 & \textbf{0.304} & \textbf{0.0020} & 32.0 & 2.145 & 0.72 & \textbf{0.333} & \textbf{0.0024} \\
\toprule
\end{tabular}
}
\end{adjustbox}
\end{adjustwidth}
\end{table}

We inspect GMPI-synthesized images and find two main reasons for our sub-optimal `Depth' scores in~\tabref{tab: qunatitative}: 1) artifacts produced by \stylegan; 2) specular reflections on images deteriorate geometry generation. We discuss both in detail below.

\begin{figure}[htpb]
    \centering
    \captionsetup[subfigure]{aboveskip=1pt}
    \begin{subfigure}{0.8\textwidth}
        \centering
        \includegraphics[width=\textwidth]{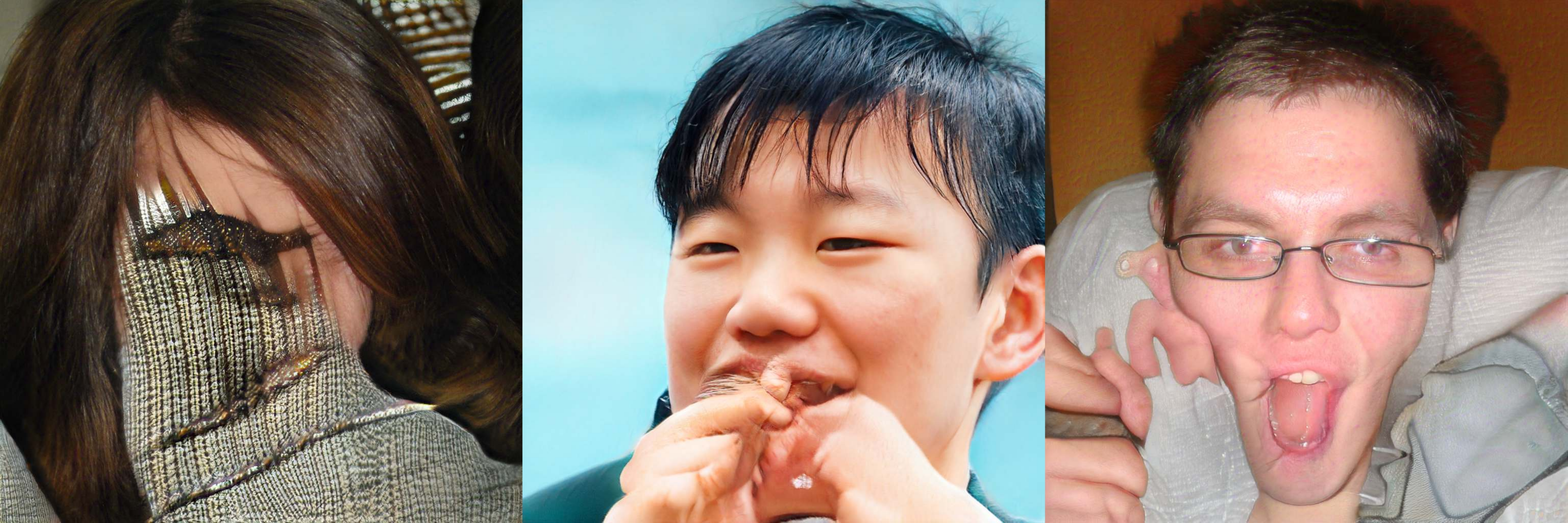}
        \captionsetup{width=\textwidth}
        \caption{$\psi = 1.0$.}
        \label{supp fig: stylegan2 psi 1.0}
    \end{subfigure}%
    \hfill
    \begin{subfigure}{0.8\textwidth}
        \centering
        \includegraphics[width=\textwidth]{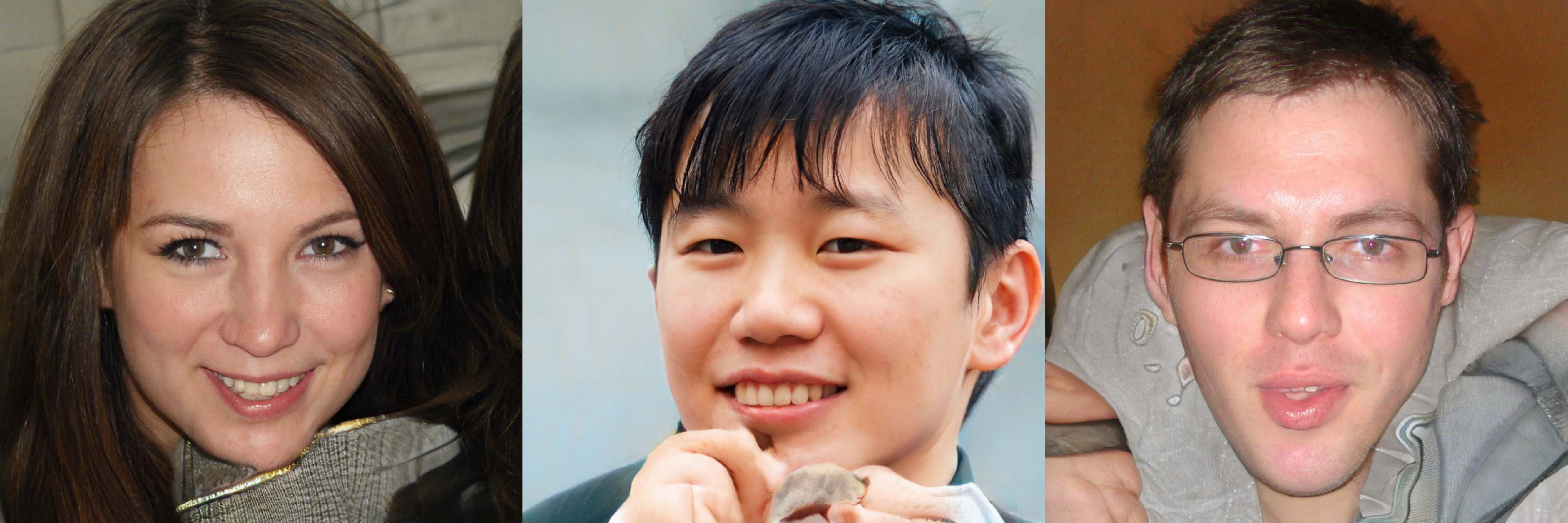}
        \captionsetup{width=\textwidth}
        \caption{$\psi = 0.7$.}
        \label{supp fig: stylegan2 psi 0.7}
    \end{subfigure}%
    \hfill
    \begin{subfigure}{0.8\textwidth}
        \centering
        \includegraphics[width=\textwidth]{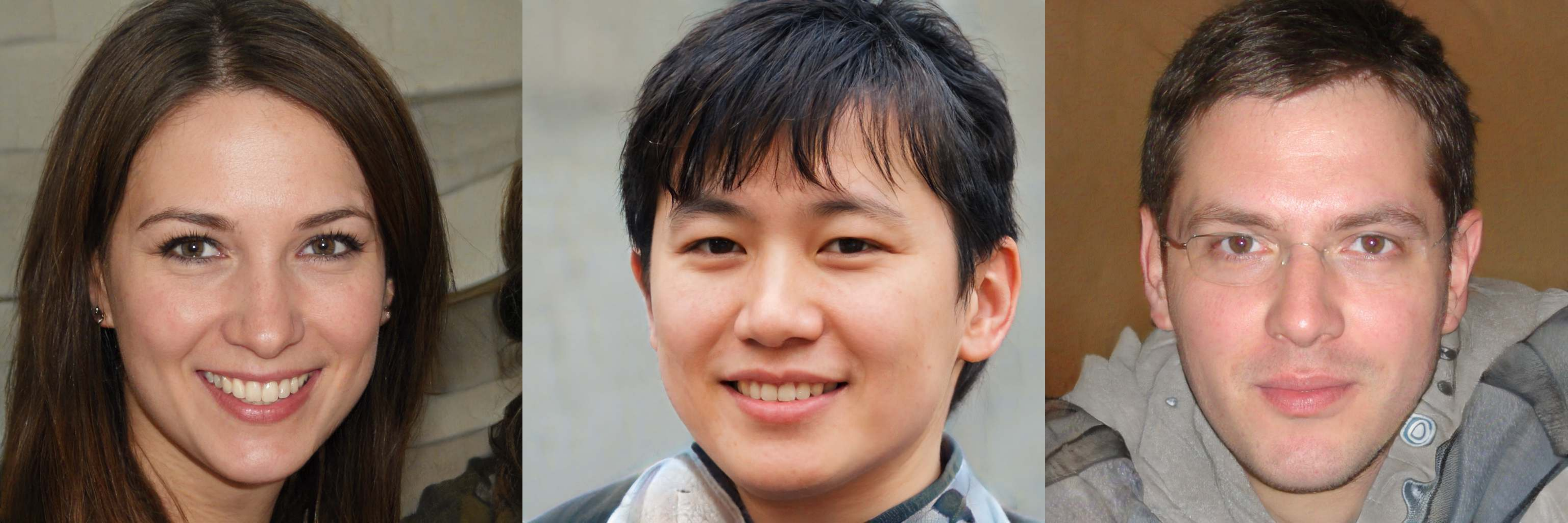}
        \captionsetup{width=\textwidth}
        \caption{$\psi = 0.5$.}
        \label{supp fig: stylegan2 psi 0.5}
    \end{subfigure}%
    \caption{
    \textbf{Effects of truncation level $\psi$ for \stylegan.}
    Images are generated with the officially-released code and checkpoints.\protect\footnotemark~
    Without truncation,~\ie,~for $\psi = 1.0$, \stylegan~generates images with significant artifacts. This degrades alpha map generation.
    The images can be generated by running the command {\tt python generate.py --outdir=out --trunc=1.0 --seeds=10,56,88 --network=https://nvlabs-fi-cdn.nvidia.com/stylegan2-ada-pytorch/\\pretrained/ffhq.pkl} while setting {\tt --trunc} to 1.0, 0.7, and 0.5 respectively.
    }
    \label{supp fig: stylegan2 psi}
\end{figure}

\begin{figure}[htpb]
    \centering
    \captionsetup[subfigure]{aboveskip=1pt}
    \begin{subfigure}{0.7\textwidth}
        \centering
        \includegraphics[width=\textwidth]{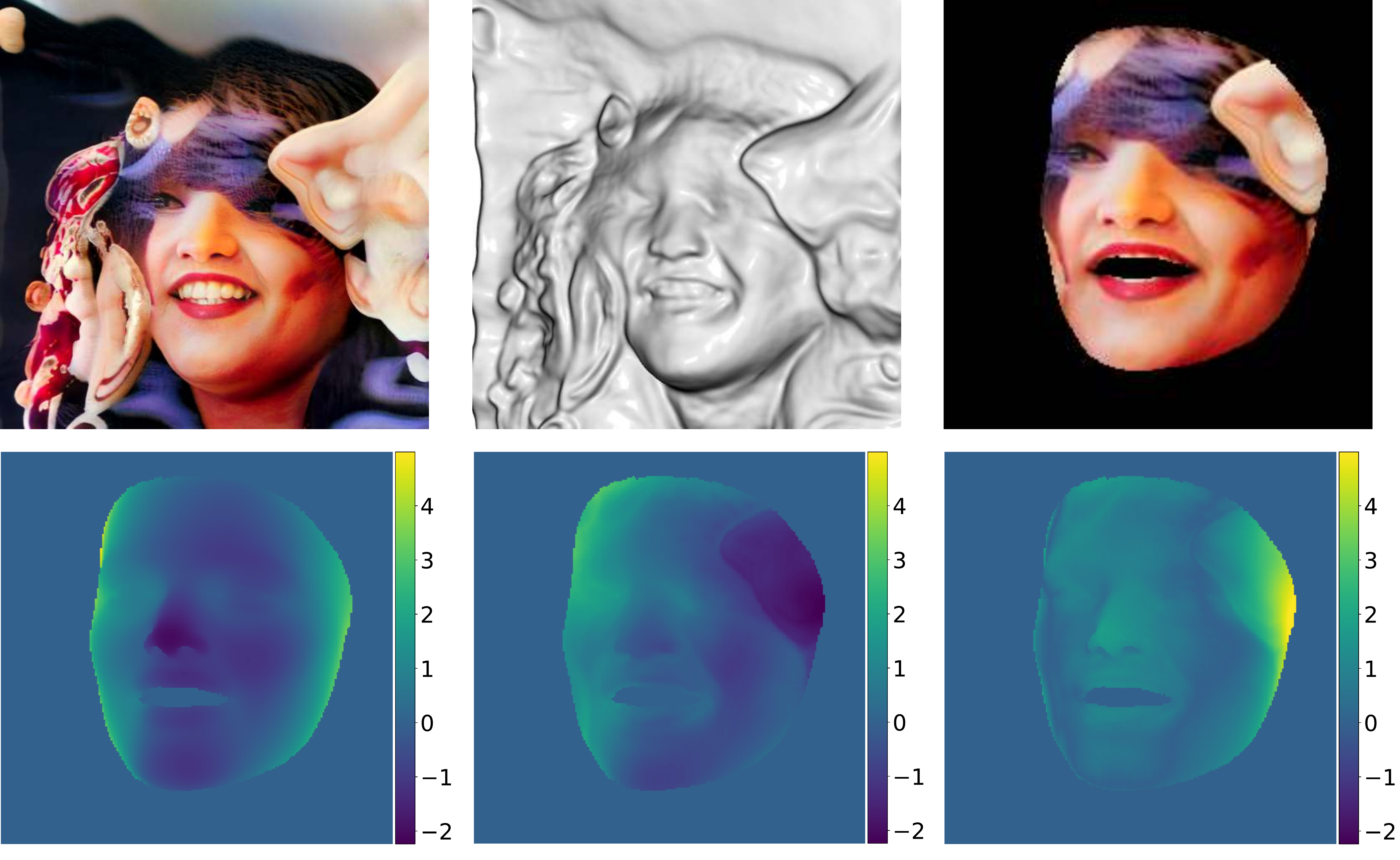}
        \captionsetup{width=\textwidth}
        \caption{$\psi = 1.0$, Depth score = 1.44.}
        \label{depth score quanlitative psi 1.0}
    \end{subfigure}%
    \hfill
    \begin{subfigure}{0.7\textwidth}
        \centering
        \includegraphics[width=\textwidth]{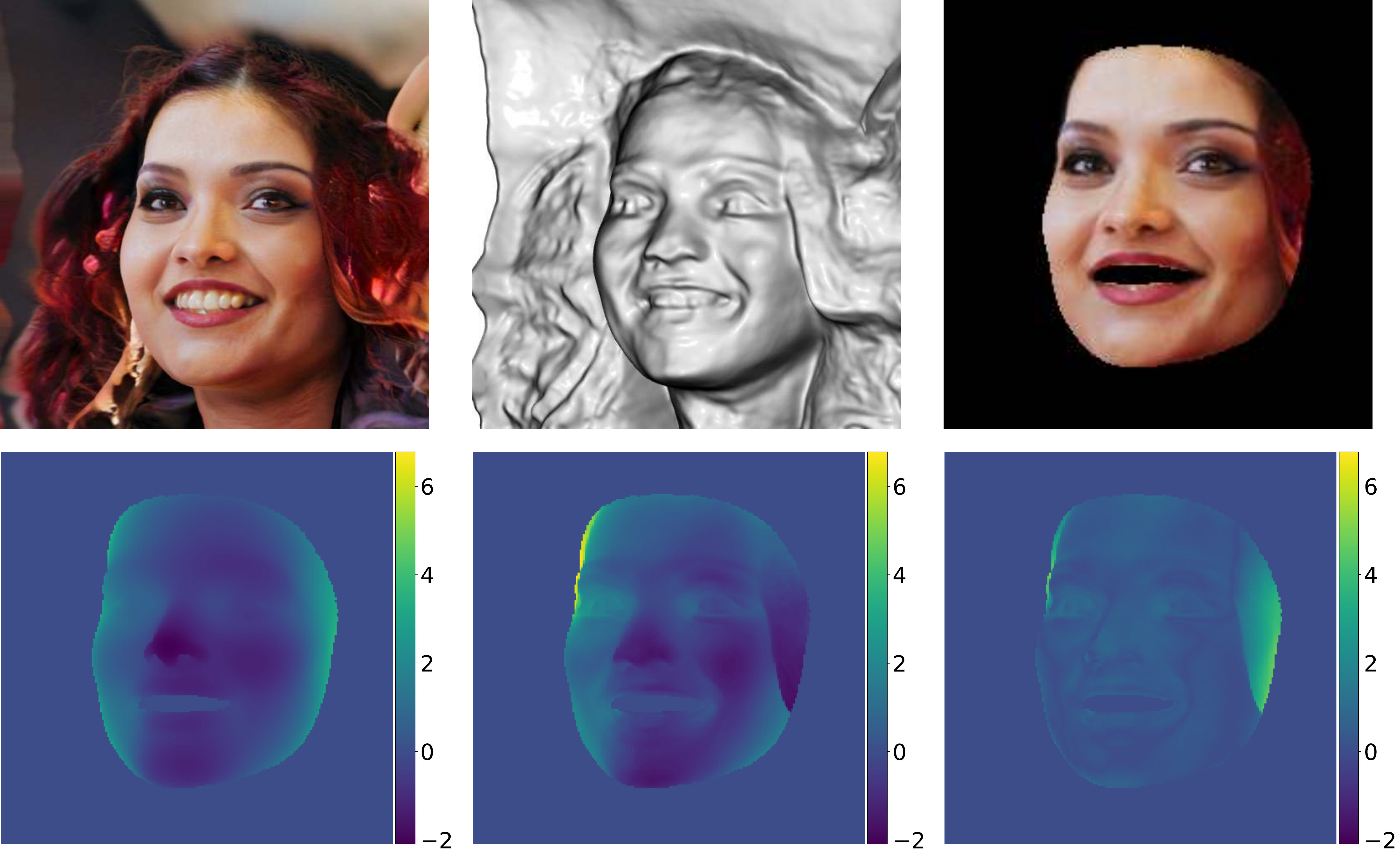}
        \captionsetup{width=\textwidth}
        \caption{$\psi = 0.5$, Depth score = 0.83. The large difference in depth prediction appearing on the cheek is primarily due to hair. The parametric facial model does not reconstruct hair and accessories such as glasses (see Fig.~2 in~\cite{Deng2019Accurate3F}).
        }
        \label{depth score quanlitative psi 0.5}
    \end{subfigure}%
    \caption{
    \textbf{Truncation level $\psi$ affects geometry generation.}
    GMPI generated scenes with the same latent variable $\bm{z}$ using two different truncation levels. Each scene is rendered with the same pose where the `Depth' score is computed.
    For each subplot, from top to bottom, left to right, we show 
    1) rendered image; 2) corresponding geometry; 3) predicted face  model~\cite{Deng2019Accurate3F};  4)  normalized pseudo ground-truth depth map; 5) normalized GMPI generated depth map (the smaller value indicates closer distance to the camera); and 6) difference between normalized depth maps 4) and 5). Smaller truncation levels $\psi$ benefit `Depth' scores at the cost of increased FID/KID values.  
    }
    \label{supp fig: depth score quanlitative}
\end{figure}

\noindent\textbf{1) Artifacts in~\stylegan.} Truncation was introduced in~\cite{Karras2019ASG,Brock2019LargeSG} to balance  between variety and fidelity. Specifically, using a truncation level $\psi \in [0, 1]$, we replace the style embedding $\bm{\omega}$ in~\equref{eq: stylegan w} with
\begin{align}
    \bm{\omega}^\prime = \bar{\bm{\omega}} + \psi \cdot (\bm{\omega} - \bar{\bm{\omega}}), %
\end{align}
where $\bar{\bm{\omega}} = \mathbb{E}_{\bm{z}} [ f_\texttt{Mapping} (\bm{z}) ]$ represents the style embedding space's center of mass. In practice, $\bar{\bm{\omega}}$ is approximated by computing the moving average of all $\bm{\omega}$ encountered during training.
Without  truncation,~\ie,~for $\psi = 1.0$, we find \stylegan~can produce results with significant artifacts as shown in \figref{supp fig: stylegan2 psi 1.0}. This demonstrates that the generator fails to convert the corresponding $\bm{\omega}$ properly. Although these artifacts are not being reflected in our 2D GAN metrics,~\ie,~FID and KID, they do affect the alpha map generation adversely. Specifically, the alpha maps are produced based on feature $\mathcal{F}_{\alpha_i}^h$, which is largely determined by the style embedding $\bm{\omega}$  through~\equref{eq: stylegan w} and~\equref{eq: F_alpha}. Consequently, artifacts cause an inferior `Depth' score.
To understand the effect of the truncation level $\psi$ during inference, we evaluate GMPI using various truncation levels $\psi$ in~\tabref{supp tab: depth score}.
As can be seen clearly, with smaller $\psi$, we consistently perform better on geometry metrics,~\ie,~lower `Depth' and `Pose' error, while trading in variety,~\ie,~higher FID/KID.
We provide a qualitative example in~\figref{supp fig: depth score quanlitative}.
Note, we do not apply  truncation  for any other quantitative  results reported in this paper.~\Ie, we always use $\psi = 1.0$ for quantitative evaluation results except in~\tabref{supp tab: depth score}.

\begin{figure}[htpb]
    \centering
    \captionsetup[subfigure]{aboveskip=1pt}
    \includegraphics[width=0.7\textwidth]{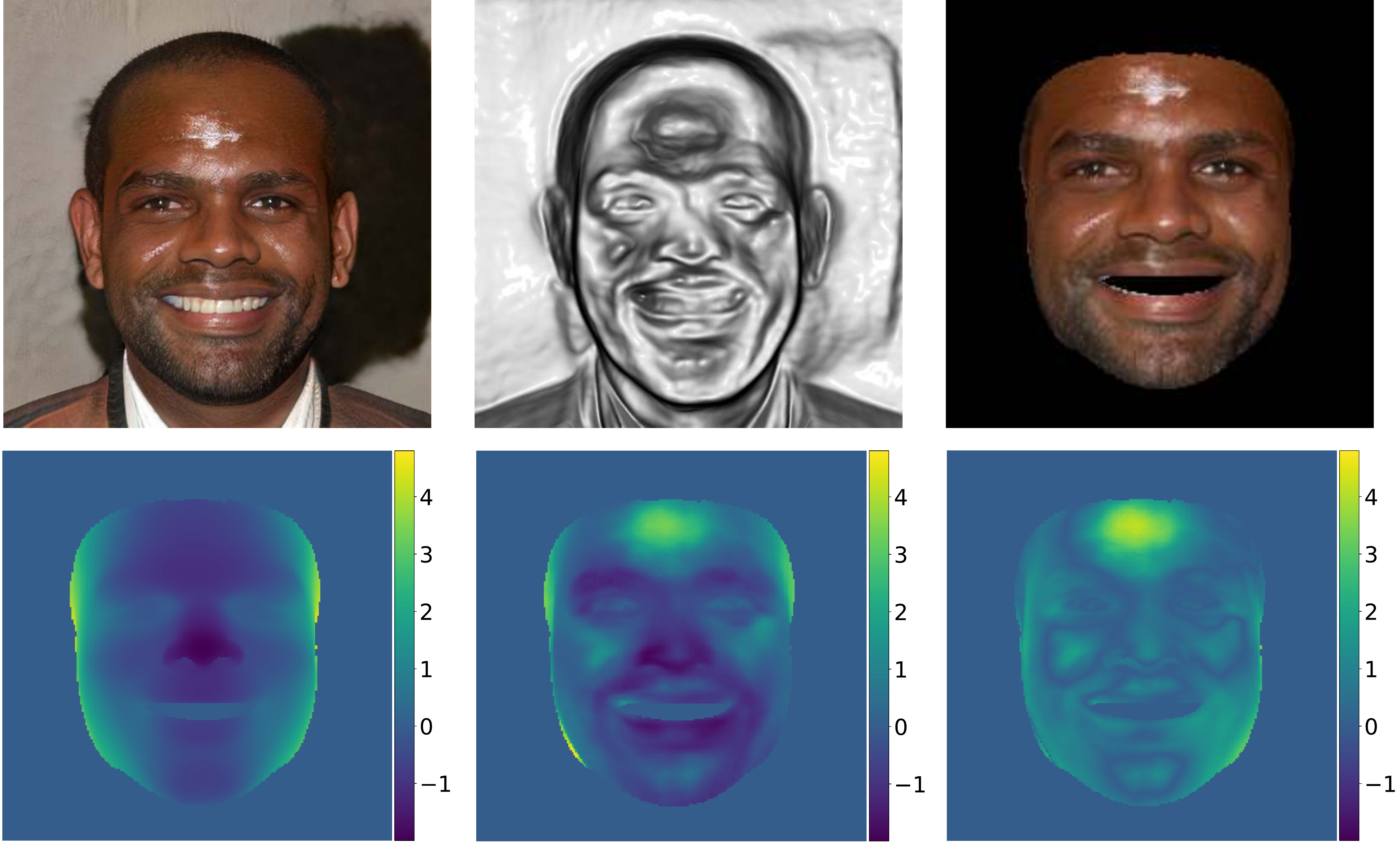}
    \caption{
    \textbf{Specular reflection deteriorates geometry generation.}
    For this example, the depth score is 1.26. The strong lighting effect on the forehead breaks the Lambertian assumption  and degrades alpha map generation.
    }
    \label{supp fig: depth score specularity}
\end{figure}

\noindent\textbf{2) Specular reflections.}
Our rendering model is not designed to handle specular reflections.  Strong specular reflections in the training data tend to degrade geometry generation, producing artifacts such as concave foreheads.
We provide a qualitative example in~\figref{supp fig: depth score specularity}.
Specular reflections are a common failure mode of geometry generation. For instance, see Sec.~5 and Fig.~8 in StyleSDF~\cite{OrEl2021StyleSDFH3}. We leave addressing of this issue to future work.

\subsection{More Ablations}\label{supp sec: more ablations}

\noindent\textbf{\#Planes During Training.} We ablate the  \texttt{\#planes} during training in~\tabref{supp tab: ablate n_planes} and~\figref{supp fig: ablate n_planes train}. We observe: the more planes we can provide during training, the better the results of GMPI.

\begin{table}[t]
\renewcommand{\arraystretch}{1.0}
\begin{adjustwidth}{0.0cm}{}
\captionsetup{width=\linewidth}
\caption{
\textbf{Ablation studies on \#planes during training.}
No truncation  is applied.
We evaluate at a resolution of $512^2$.
\texttt{DPC} refers to discriminator pose conditioning (\secref{sec:dpc}), 
$\mathcal{F}_{\alpha_i}^h$ refers to the plane-specific feature introduced in~\equref{eq: F_alpha}, and 
\texttt{Shading} indicates the shading-guided training discussed in~\secref{sec:ma}.
\texttt{\#planes} denotes the number of planes we used during training.
During inference, we use 32 planes.
We observe, the more planes we  provide during training, the better the results of GMPI.
Please see~\figref{supp fig: ablate n_planes train} for qualitative examples.
}
\label{supp tab: ablate n_planes}
\renewcommand\theadfont{}
\centering
\setlength\aboverulesep{0pt}
\setlength\belowrulesep{0pt}
\setlength{\tabcolsep}{3pt}
{
\small
\begin{tabular}{lccccc@{\hskip 0.3em}r@{\hskip 0.65em}r@{\hskip 0.65em}r@{\hskip 0.65em}r@{\hskip 0.65em}r@{\hskip 0.65em}r@{\hskip 0.65em}r} 
\toprule
& \multirow{2}{*}{\texttt{\#planes}} & \multirow{2}{*}{\texttt{DPC}} & \multirow{2}{*}{$\mathcal{F}_{\alpha_i}^h$} & \multirow{2}{*}{\texttt{Shading}} & \multicolumn{5}{c}{FFHQ} & \multicolumn{2}{c}{\afhq-Cat} \\
\cmidrule(r){6-10} \cmidrule(){11-12}
 & & & & & FID$\downarrow$ & KID$\downarrow$  & ID$\uparrow$ & Depth$\downarrow$ & Pose$\downarrow$ & FID$\downarrow$ & KID$\downarrow$ \\
\midrule
(a) & 32 & \cmark & \cmark & \cmark & \textbf{7.40} & \textbf{0.337} & \textbf{0.74} & \textbf{0.457} & \textbf{0.006} & 7.93 & 0.489 \\
(b) & 16 & \cmark & \cmark & \cmark &  9.83 & 0.575 & 0.74 & 0.574 & 0.007 & \textbf{6.82} & \textbf{0.358} \\
(c) & 8 & \cmark & \cmark & \cmark & 11.4 & 0.657 & 0.72 & 0.778 & 0.008 & 7.26 & 0.384 \\
(d) & 4 & \cmark & \cmark & \cmark & 16.4 & 1.043 & 0.65 & 0.992 & 0.007 & 8.53 & 0.467 \\
\toprule
\end{tabular}
}
\end{adjustwidth}
\end{table}

\begin{figure}[!t]
    \centering
    \captionsetup[subfigure]{aboveskip=1pt}
    \begin{subfigure}{0.19\textwidth}
        \centering
        \includegraphics[width=\textwidth]{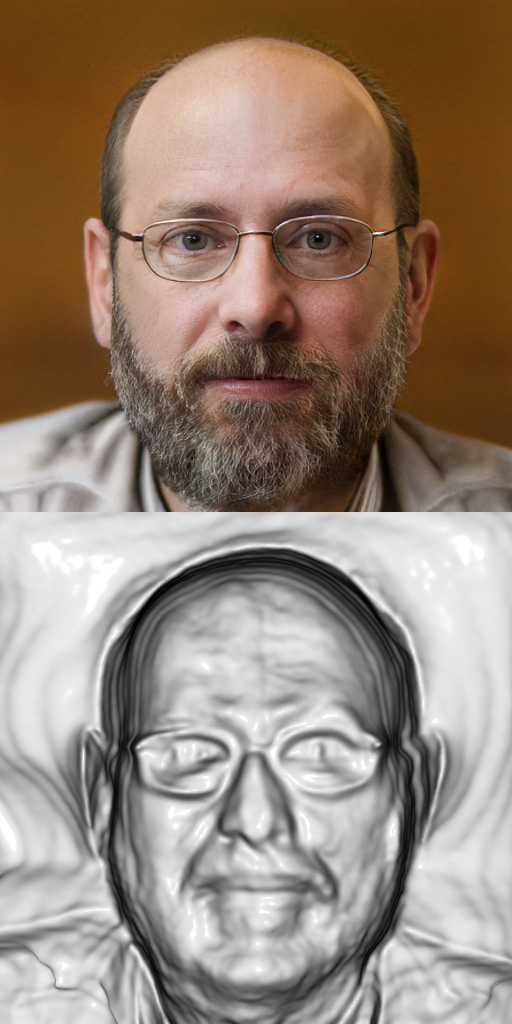}
        \captionsetup{width=\textwidth}
        \caption{$0\sigma$.}
        \label{supp fig: cam noise 0}
    \end{subfigure}%
    \hfill
    \begin{subfigure}{0.19\textwidth}
        \centering
        \includegraphics[width=\textwidth]{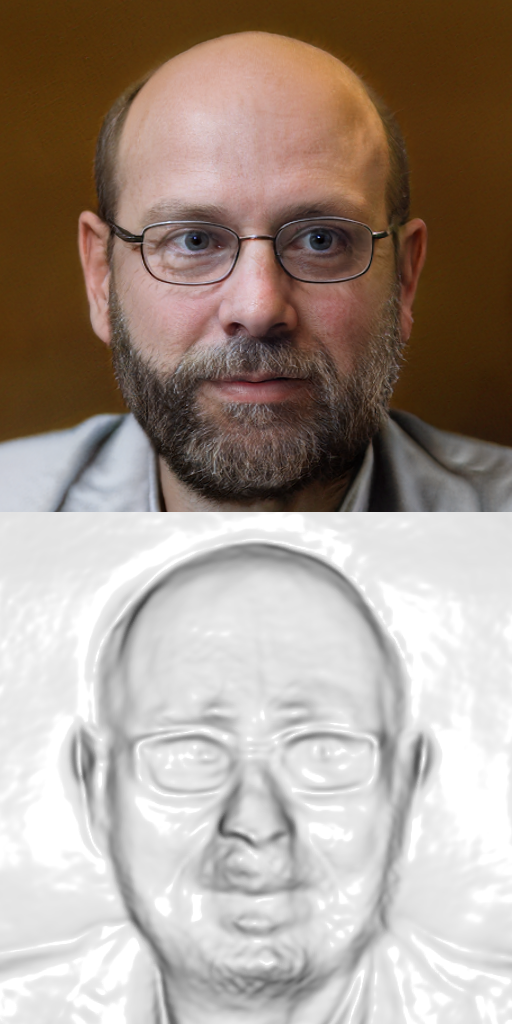}
        \captionsetup{width=\textwidth}
        \caption{$1\sigma$.}
        \label{supp fig: cam noise 1}
    \end{subfigure}%
    \hfill
    \begin{subfigure}{0.19\textwidth}
        \centering
        \includegraphics[width=\textwidth]{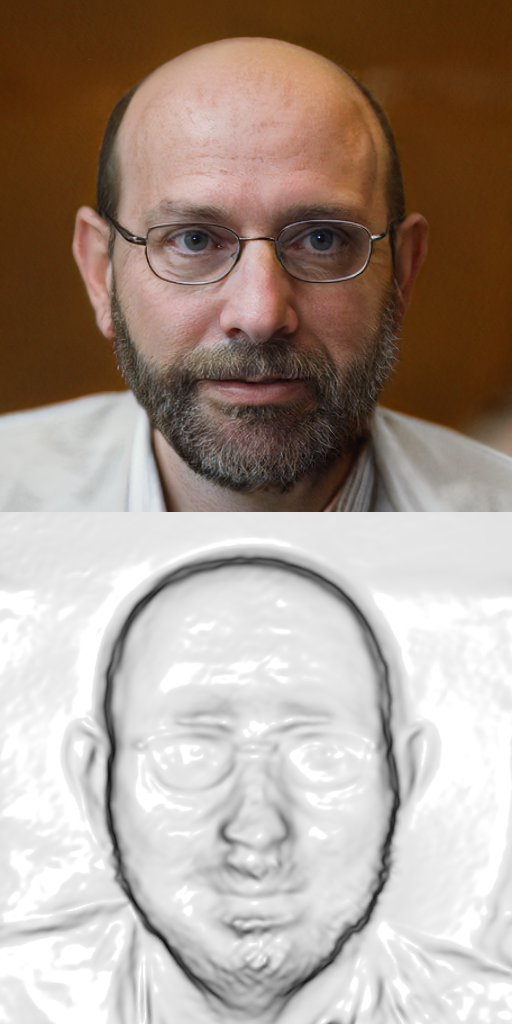}
        \captionsetup{width=\textwidth}
        \caption{$2\sigma$.}
        \label{supp fig: cam noise 2}
    \end{subfigure}%
    \hfill
    \begin{subfigure}{0.19\textwidth}
        \centering
        \includegraphics[width=\textwidth]{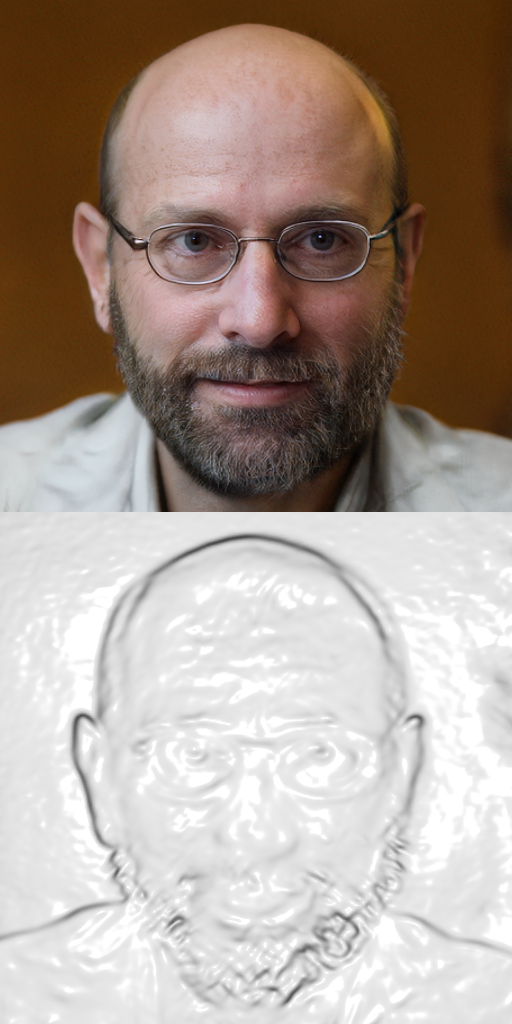}
        \captionsetup{width=\textwidth}
        \caption{$3\sigma$.}
        \label{supp fig: cam noise 3}
    \end{subfigure}%
    \hfill
    \begin{subfigure}{0.19\textwidth}
        \centering
        \includegraphics[width=\textwidth]{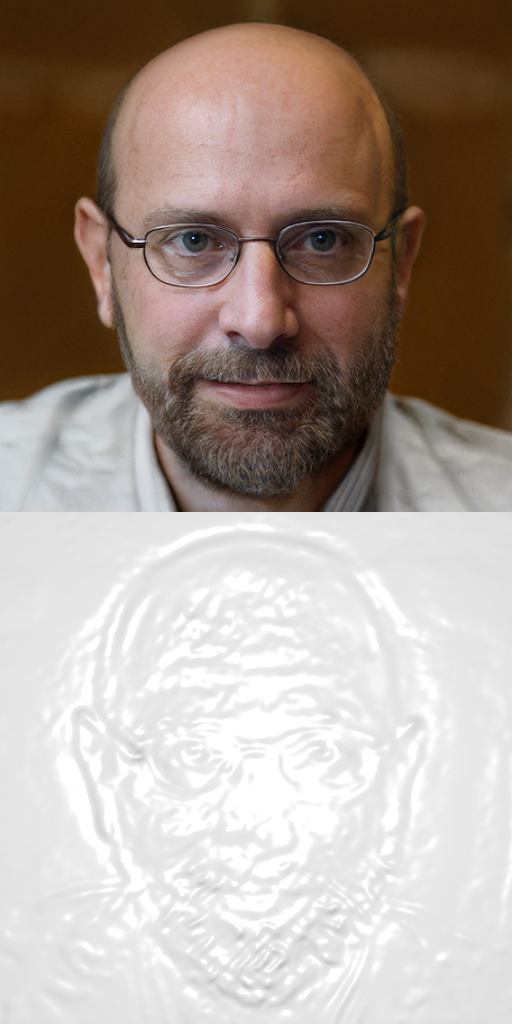}
        \captionsetup{width=\textwidth}
        \caption{$4\sigma$.}
        \label{supp fig: cam noise 4}
    \end{subfigure}%
    \caption{
    \textbf{Ablate robustness to inaccurate camera poses.} Panels (a)-(e) correspond to \tabref{supp tab: ablate cam noises}'s rows (a)-(e). All results are generated from the same latent code $\bm{z}$. Within expectation, the more inaccurate camera poses the model is trained with, the less 3D geometry we can obtain.
    }
    \label{supp fig: ablate cam noises}
\end{figure}

\begin{figure}[!t]
    \centering
    \captionsetup[subfigure]{aboveskip=1pt}
    \begin{subfigure}{\textwidth}
        \centering
        \includegraphics[width=\textwidth]{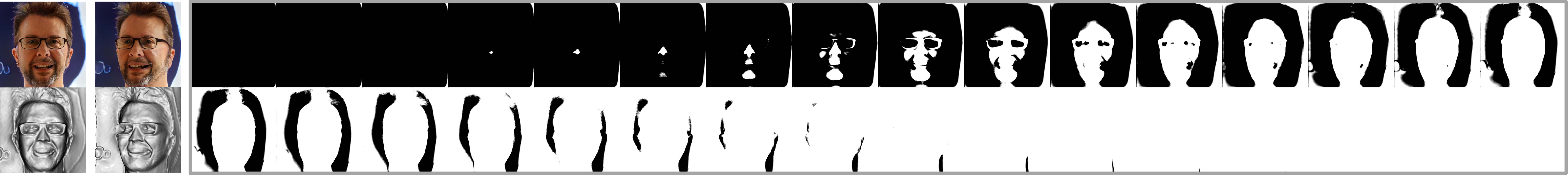}
        \captionsetup{width=\textwidth}
        \caption{32 planes.}
        \label{supp fig: 32 planes train}
    \end{subfigure}%
    \hfill
    \begin{subfigure}{\textwidth}
        \centering
        \includegraphics[width=\textwidth]{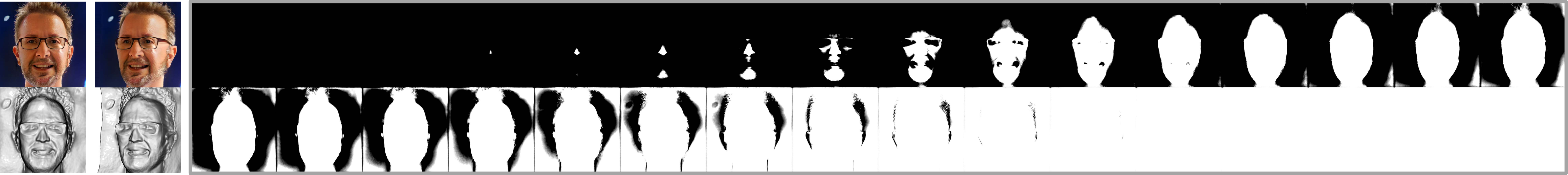}
        \captionsetup{width=\textwidth}
        \caption{16 planes.}
        \label{supp fig: 16 planes train}
    \end{subfigure}%
    \hfill
    \begin{subfigure}{\textwidth}
        \centering
        \includegraphics[width=\textwidth]{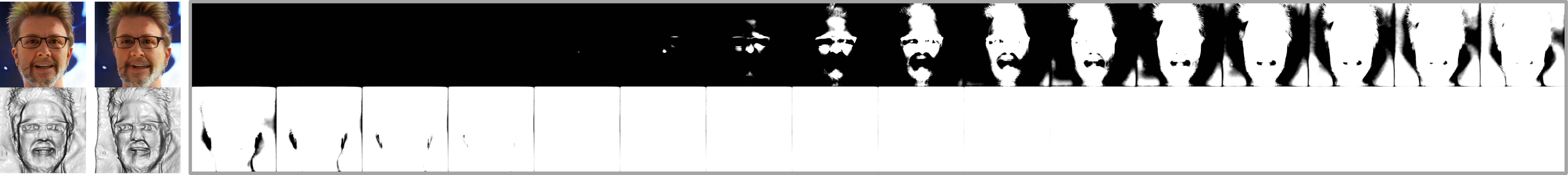}
        \captionsetup{width=\textwidth}
        \caption{8 planes.}
        \label{supp fig: 8 planes train}
    \end{subfigure}%
    \hfill
    \begin{subfigure}{\textwidth}
        \centering
        \includegraphics[width=\textwidth]{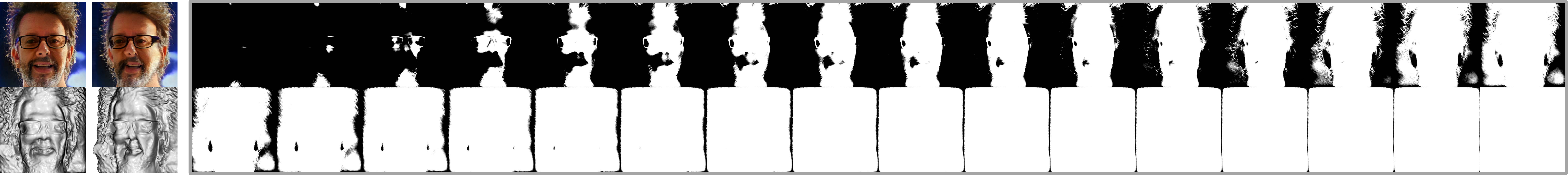}
        \captionsetup{width=\textwidth}
        \caption{4 planes.}
        \label{supp fig: 4 planes train}
    \end{subfigure}%
    \caption{
    \textbf{Ablate \#planes during training.}
    For each subplot, from left to right, we show two views and the 32 alpha maps that GMPI produces during inference. Each subplot's caption indicates the number of planes used during training.
    We observe: the more planes we can provide during training, the better the results of GMPI.
    }
    \label{supp fig: ablate n_planes train}
\end{figure}

\noindent\textbf{Robustness to inaccurate camera poses.}
To understand the robustness to inaccurate camera poses, we add noise to the observer's camera pose. Specifically, we follow~\cite{ChanARXIV2021} to first compute per-element standard deviation $\sigma$ of the estimated camera pose matrices for real images. During training, we add noise of  $\{1\sigma, 2\sigma, 3\sigma, 4\sigma\}$  to each element of the camera pose matrix.
We report results in~\tabref{supp tab: ablate cam noises} and~\figref{supp fig: ablate cam noises}.
We want to emphasize, the estimated camera poses for real images are not perfect. Therefore, $0\sigma$ does not indicate fully-accurate camera pose information.~\figref{supp fig: ablate cam noises} verifies that the more noise we add, the less photorealistic the geometry we obtain. This is aligned with the trend of the Depth and Pose metrics in~\tabref{supp tab: ablate cam noises}. Note, FID, KID, and ID metrics are misleading as we do not observe much difference. This verifies again that 2D metrics lack the ability to capture 3D errors.

\begin{table}[t]
\renewcommand{\arraystretch}{1.0}
\begin{adjustwidth}{0.0cm}{}
\captionsetup{width=\linewidth}
\caption{
\textbf{Ablation study on camera pose accuracy.}
No truncation  is applied.
We evaluate at a resolution of $512^2$.
\texttt{Noise} indicates the level of noise we added to the camera poses during training. Specifically, $1\sigma$ denotes camera poses are corrupted with one standard deviation of the estimated camera pose matrices for images from the corresponding dataset (see~\secref{supp sec: more ablations}).
All results are trained with \texttt{DPC}, plane-specific features $\mathcal{F}_{\alpha_i}^h$, and shading-guided training with 32 planes during training.
}
\label{supp tab: ablate cam noises}
\renewcommand\theadfont{}
\centering
\setlength\aboverulesep{0pt}
\setlength\belowrulesep{0pt}
\setlength{\tabcolsep}{3pt}
{
\small
\begin{tabular}{lc@{\hskip 0.3em}r@{\hskip 0.65em}r@{\hskip 0.65em}r@{\hskip 0.65em}r@{\hskip 0.65em}r@{\hskip 0.65em}r@{\hskip 0.65em}r} 
\toprule
& \multirow{2}{*}{\texttt{Noise} } & \multicolumn{5}{c}{FFHQ} & \multicolumn{2}{c}{\afhq-Cat} \\
\cmidrule(r){3-7} \cmidrule(){8-9}
 & & FID$\downarrow$ & KID$\downarrow$  & ID$\uparrow$ & Depth$\downarrow$ & Pose$\downarrow$ & FID$\downarrow$ & KID$\downarrow$ \\
\midrule
(a) & $0 \sigma$ & 8.29 & 0.454 & 0.74 & 0.46 & 0.006 & 7.79 & 0.474  \\

(b) & $1 \sigma$ & 6.02 & 0.269 & 0.81 & 0.81 & 0.017 & 5.95 & 0.296 \\
(c) & $2 \sigma$ & 5.38 & 0.221 & 0.86 & 1.00 & 0.027 & 5.51 & 0.266 \\
(d) & $3 \sigma$ & 5.26 & 0.191 & 0.89 & 1.37 & 0.038 & 7.05 & 0.397 \\
(e) & $4 \sigma$ & 4.72 & 0.176 & 0.90 & 1.68 & 0.046 & 6.92 & 0.393 \\
\toprule
\end{tabular}
}
\end{adjustwidth}
\end{table}

\footnotetext{\url{https://github.com/NVlabs/stylegan2-ada-pytorch}}

\section{Implementation Details}\label{supp sec: implement}

\noindent\textbf{Hyperparameters.} \tabref{supp tab: implement} summarizes the hyperparameters we used for training GMPI. We perform all training on 8 Tesla V100 GPUs using 5000 iterations. For all experiments except FFHQ256, we use a batch size of 32. This exposes the discriminator  to 0.16 million real images. FFHQ256 is trained on 0.32 million real images since it uses a larger batch size of 64.
We apply the minibatch standard deviation layer~\cite{Karras2018ProgressiveGO} independently on each image.
We apply $x$-flip data augmentation on \afhq~and MetFaces.
To determine the channel number $\texttt{dim}_h$ of $\mathcal{F}^h$ (\equref{eq: stylegan C^h}) and $\mathcal{F}^h_{\alpha_i}$ (\equref{eq: F_alpha}), we follow~\cite{Karras2020TrainingGA}'s official implementation\footnote{\url{https://github.com/NVlabs/stylegan2-ada-pytorch}} using: 
\begin{align}
    \texttt{dim}_h = \min (\frac{2^{15}}{h}, 512),
\end{align}
where $2^{15}$ is the channel base number. The only exception is FFHQ256, where we use a channel base number of $2^{14}$ following the official implementation.
All experiments utilize R1 penalties~\cite{Mescheder2018WhichTM} with weight 10.0 and learning rate $2\times 10^{-3}$. 
We use Adam~\cite{Kingma2015AdamAM} as the optimizer.
In all experiments, we use 32 planes during training.

\noindent\textbf{Mixed precision.} We follow \stylegan's official code and use half precision for both generator and discriminator layers corresponding to the four highest resolutions.

\noindent\textbf{Near/far depths.} We set the near and far depths of MPI for each dataset as  0.95/1.12 (FFHQ and MetFaces), 2.55/2.8 (\afhq-Cat).

\noindent\textbf{Depth normalization.} The depth $d_i$ $\forall i$ is normalized to the range $[0,1]$ before being used in the embedding function $f_{\texttt{Embed}}$ in~\equref{eq: F_alpha}. Specifically, we use $d_i^\prime = \frac{d_i - d_1}{d_L - d_1}$.

\noindent\textbf{Shading-guided training.} As mentioned in~\secref{sec:ma}, we utilize~\equref{eq: shaing} to apply shading. For the first 1000 iterations, we set $k_a = 1.0$ and $k_d = 0.0$. We linearly reduce $k_a$ to 0.9 and linearly increase $k_d$ to 0.1 during iteration 1001 to 2000. Starting from the 2001$^\text{st}$ iteration, we fix $k_a = 0.9$ and $k_d = 0.1$.
Meanwhile, lighting direction $\bm{l}$ is randomly sampled. We follow~\cite{Pan2021ASG} to use $l_h \sim \mathcal{N}(0.0, 0.2)$ and $l_v \sim \mathcal{N}(0.2, 0.05)$, where $l_h$ and $l_v$ represent horizontal and vertical angles for lighting directions respectively.  

\noindent\textbf{Structure of $f_\texttt{Embed}$ (\equref{eq: F_alpha}).} For each resolution $h\times h$, we utilize three modulated-convolutional layers~\cite{Karras2020AnalyzingAI} to construct $f_\texttt{Embed}$, whose channel numbers are  $\texttt{dim}_{h / 4}$, $\texttt{dim}_{h / 2}$, $\texttt{dim}_h$ respectively.

\noindent\textbf{Alternative representation of $f_\texttt{Embed}$ (\equref{eq: F_alpha}).} Besides conditioning $f_\texttt{Embed}$ on specific depth $d_i$, we also studied use of a learnable token. Concretely,  $f_\texttt{Embed}$ was represented using a tensor of  shape $h \times h \times \texttt{dim}_h$.  This resulted in reasonable geometry but we find conditioning of alpha maps  on depth to provide better results. We leave further exploration of  learnable tokens to future work.

\noindent\textbf{Background plane.} We treat the last plane of the MPI-like representation,~\ie, the $L^{\text{th}}$ plane, as the background plane. To color this plane, we treat the left-most and right-most $5\%$ pixels of the synthesized image $C$ as the left and right boundary. RGB values of all remaining pixels on this plane are linearly interpolated between the left and the right boundary. We find this simple procedure to work well in our experiments.

\noindent\textbf{Mesh generation.} We utilize the marching cube algorithm~\cite{lorensen1987marching} implemented in PyMCubes for generating meshes.\footnote{\url{https://github.com/pmneila/PyMCubes}} We also utilize its smoothing function for better visualization.

\begin{table}[t]
\renewcommand{\arraystretch}{1.0}
\begin{adjustwidth}{0.0cm}{}
\captionsetup{width=\linewidth}
\caption{
\textbf{Hyperparameters used for training GMPI.}
}
\label{supp tab: implement}
\renewcommand\theadfont{}
\centering
\setlength\aboverulesep{0pt}
\setlength\belowrulesep{0pt}
\setlength{\tabcolsep}{3pt}
{
\small
\begin{tabular}{lccccc} 
\toprule
& FFHQ256 & FFHQ512 & FFHQ1024 & \afhq & MetFaces \\
\midrule
Resolution & $256^2$ & $512^2$ & $1024^2$ & $512^2$ & $1024^2$\\
\#GPUs   & 8 & 8 & 8 & 8 & 8 \\
Training length (iters) & 5k & 5k & 5k & 5k & 5k \\
Training length (\#imgs) & 0.32M & 0.16M & 0.16M & 0.16M & 0.16M \\
Batch size & 64 & 32 & 32 & 32 & 32 \\
Minibatch stddev~\cite{Karras2018ProgressiveGO} & 1 & 1 & 1 & 1 & 1 \\
Dataset $x$-flips & \xmark & \xmark & \xmark & \cmark & \cmark \\
\midrule
Channel base & $\frac{1}{2}\times$ & $1\times$ & $1\times$ & $1\times$ & $1\times$ \\
Learning rate ($\times 10^{-3}$) & 2 & 2 & 2 & 2 & 2 \\
R1 penalty weight~\cite{Mescheder2018WhichTM} & 10 & 10 & 10 & 10 & 10 \\
Mixed-precision & \cmark & \cmark & \cmark & \cmark & \cmark \\
\toprule
\end{tabular}
}
\end{adjustwidth}
\end{table}

\section{Additional Qualitative Results}\label{supp sec: qualitative}

\subsection{Uncurated Results}

\begin{figure}[!t]
    \centering
    \captionsetup[subfigure]{aboveskip=1pt}
    \begin{subfigure}{0.9\textwidth}
        \centering
        \includegraphics[width=\textwidth]{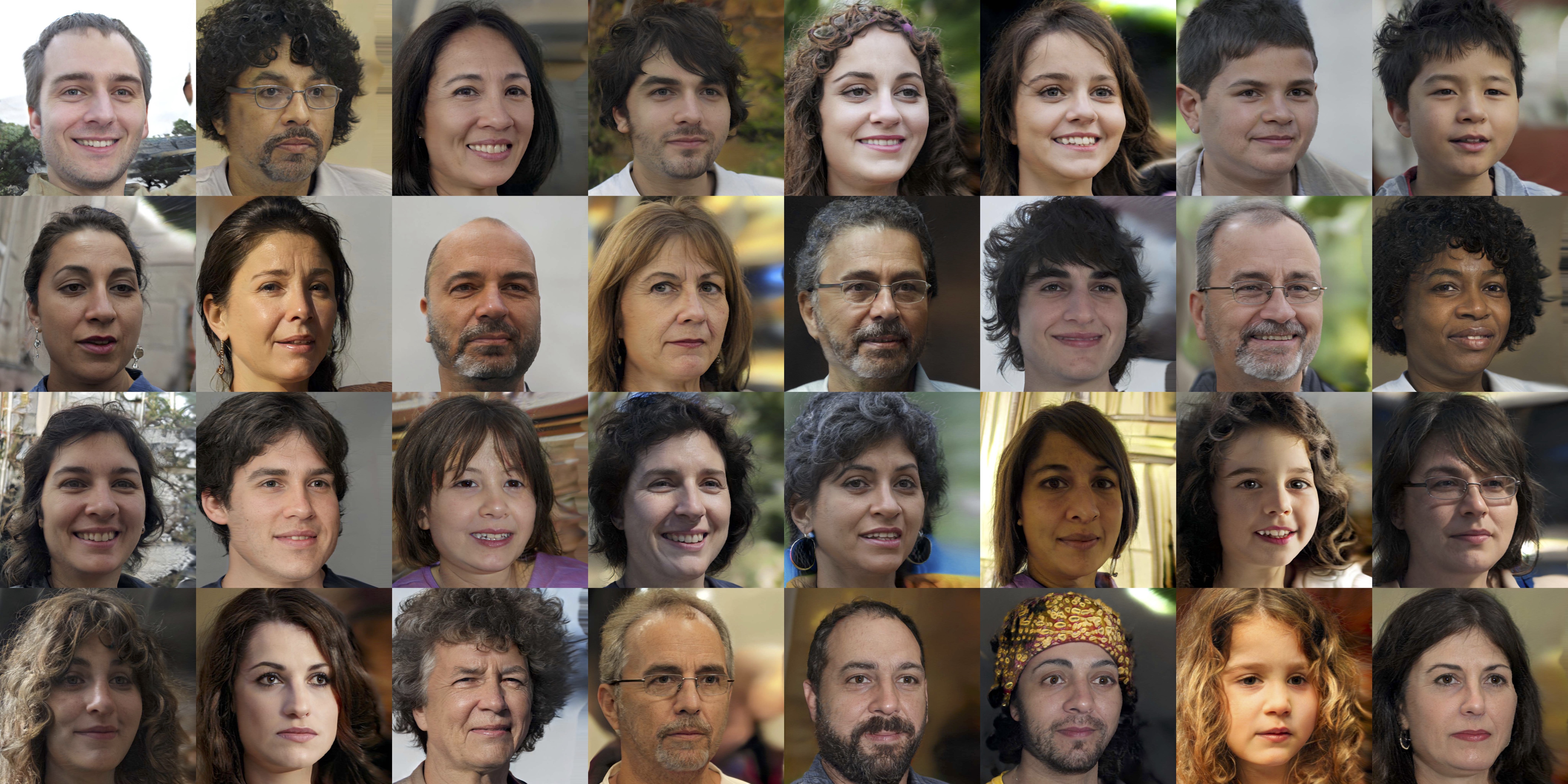}
        \captionsetup{width=\textwidth}
        \caption{Renderings for corresponding geometries in~\figref{supp fig: uncurated ffhq, mesh}.
        }
        \label{supp fig: uncurated ffhq, rgb}
    \end{subfigure}%
    \hfill
    \begin{subfigure}{0.9\textwidth}
        \centering
        \includegraphics[width=\textwidth]{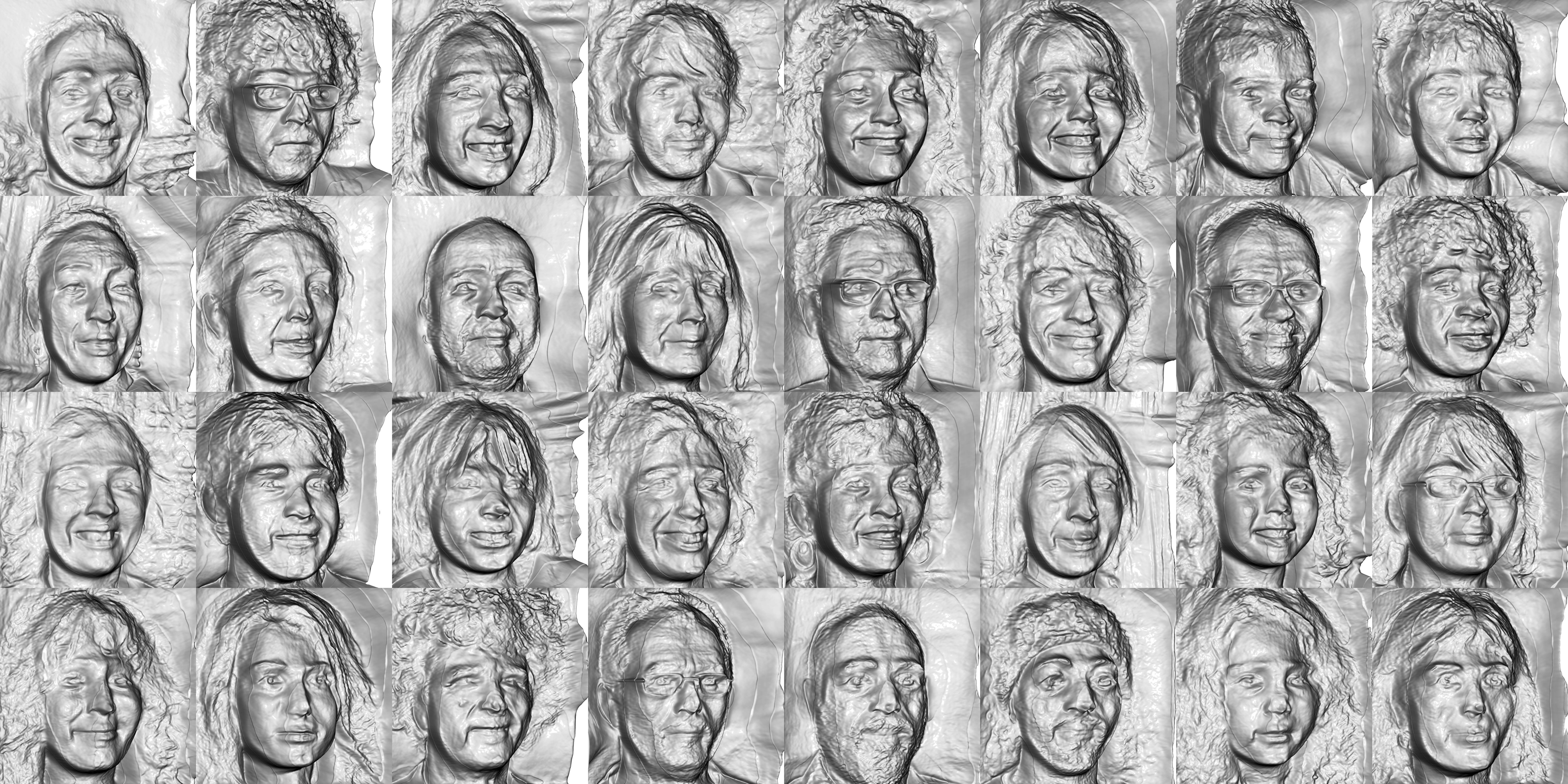}
        \captionsetup{width=\textwidth}
        \caption{Geometries for corresponding renderings in~\figref{supp fig: uncurated ffhq, rgb}.
        }
        \label{supp fig: uncurated ffhq, mesh}
    \end{subfigure}%
    \caption{
    \textbf{Uncurated results on FFHQ.} From top to bottom, left to right, we show generations with seed 1-32. Results are generated with truncation $\psi = 0.5$~\cite{Karras2019ASG}.
    }
    \label{supp fig: uncurated ffhq}
\end{figure}

\begin{figure}[!t]
    \centering
    \captionsetup[subfigure]{aboveskip=1pt}
    \begin{subfigure}{0.9\textwidth}
        \centering
        \includegraphics[width=\textwidth]{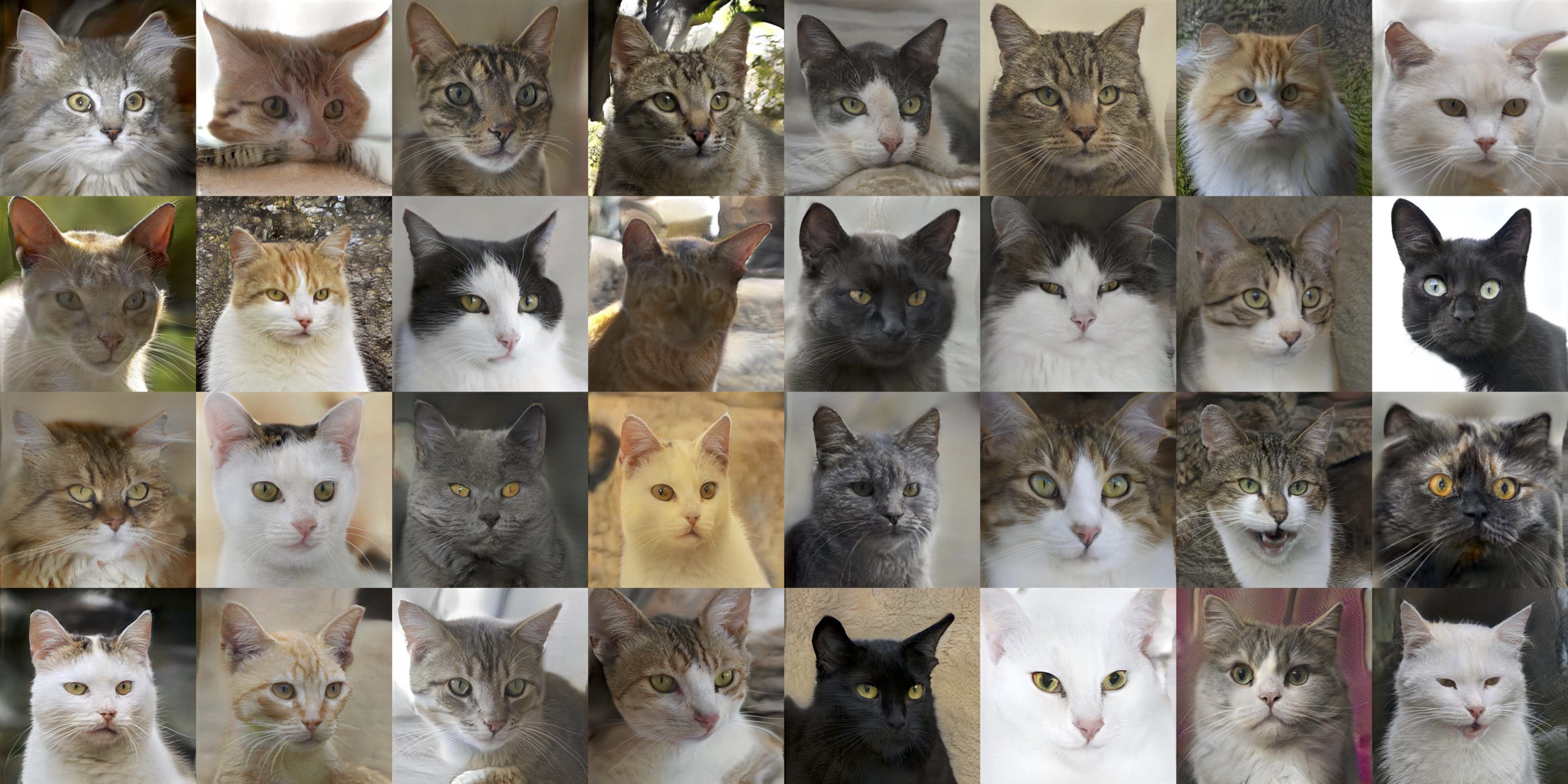}
        \captionsetup{width=\textwidth}
        \caption{Renderings for corresponding geometries in~\figref{supp fig: uncurated afhq, mesh}.
        }
        \label{supp fig: uncurated afhq, rgb}
    \end{subfigure}%
    \hfill
    \begin{subfigure}{0.9\textwidth}
        \centering
        \includegraphics[width=\textwidth]{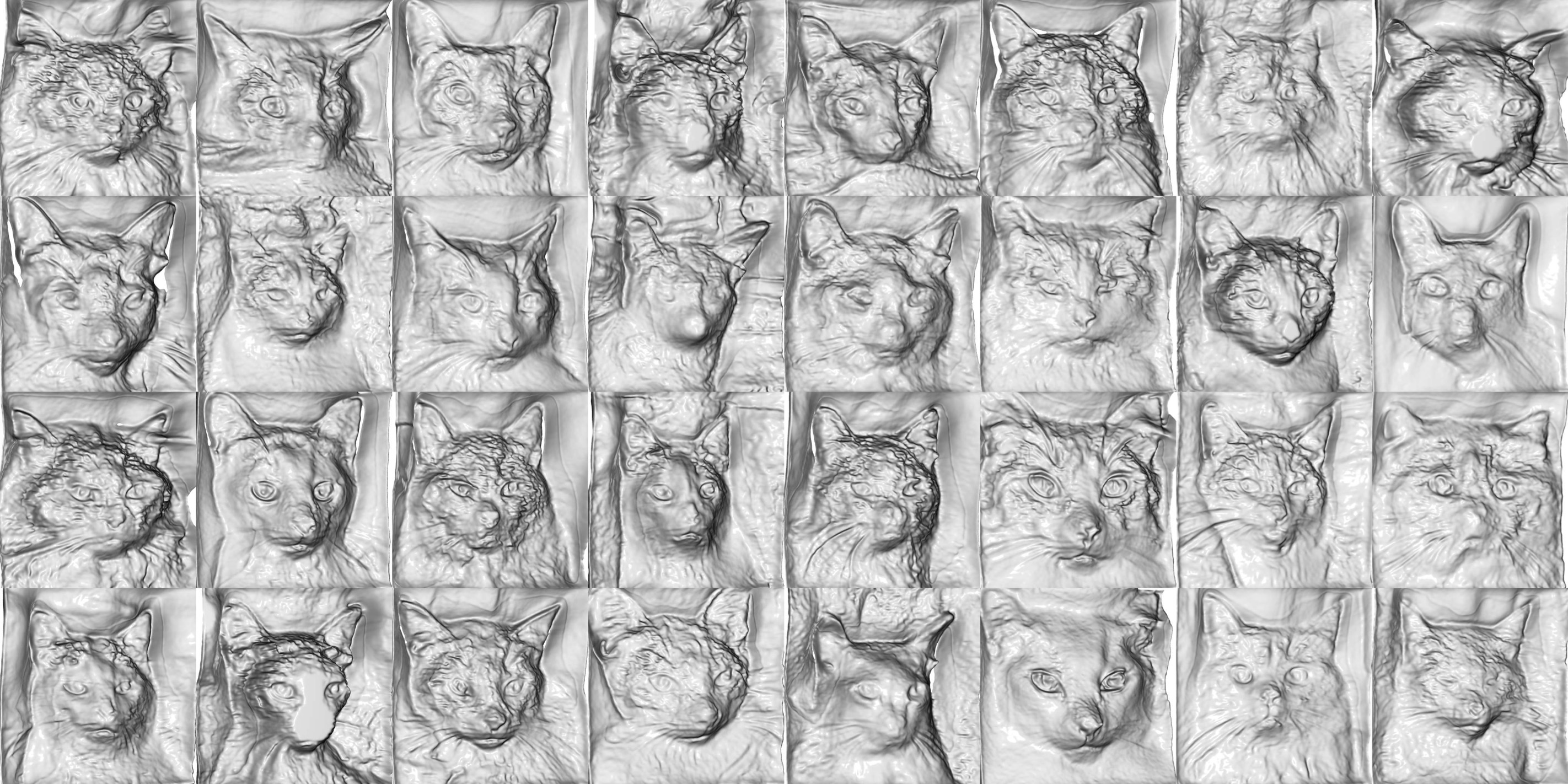}
        \captionsetup{width=\textwidth}
        \caption{Geometries for corresponding renderings in~\figref{supp fig: uncurated afhq, rgb}.
        }
        \label{supp fig: uncurated afhq, mesh}
    \end{subfigure}%
    \caption{
    \textbf{Uncurated results on \afhq.} From top to bottom, left to right, we show generations with seed 1-32. Results are generated with truncation $\psi = 0.7$~\cite{Karras2019ASG}.
    }
    \label{supp fig: uncurated afhq}
\end{figure}

\begin{figure}[!t]
    \centering
    \captionsetup[subfigure]{aboveskip=1pt}
    \begin{subfigure}{0.9\textwidth}
        \centering
        \includegraphics[width=\textwidth]{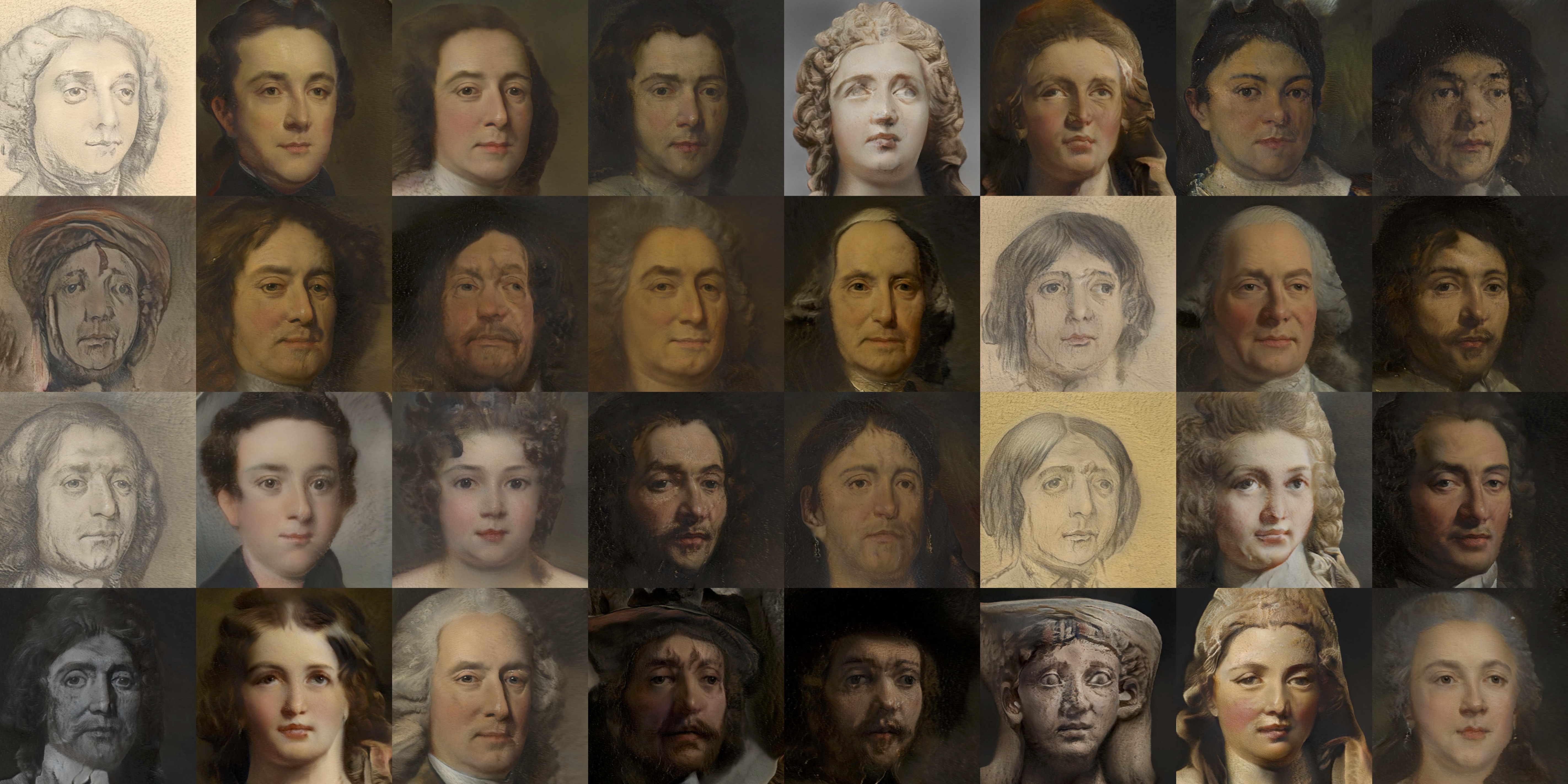}
        \captionsetup{width=\textwidth}
        \caption{Renderings for corresponding geometries in~\figref{supp fig: uncurated metfaces, mesh}.
        }
        \label{supp fig: uncurated metfaces, rgb}
    \end{subfigure}%
    \hfill
    \begin{subfigure}{0.9\textwidth}
        \centering
        \includegraphics[width=\textwidth]{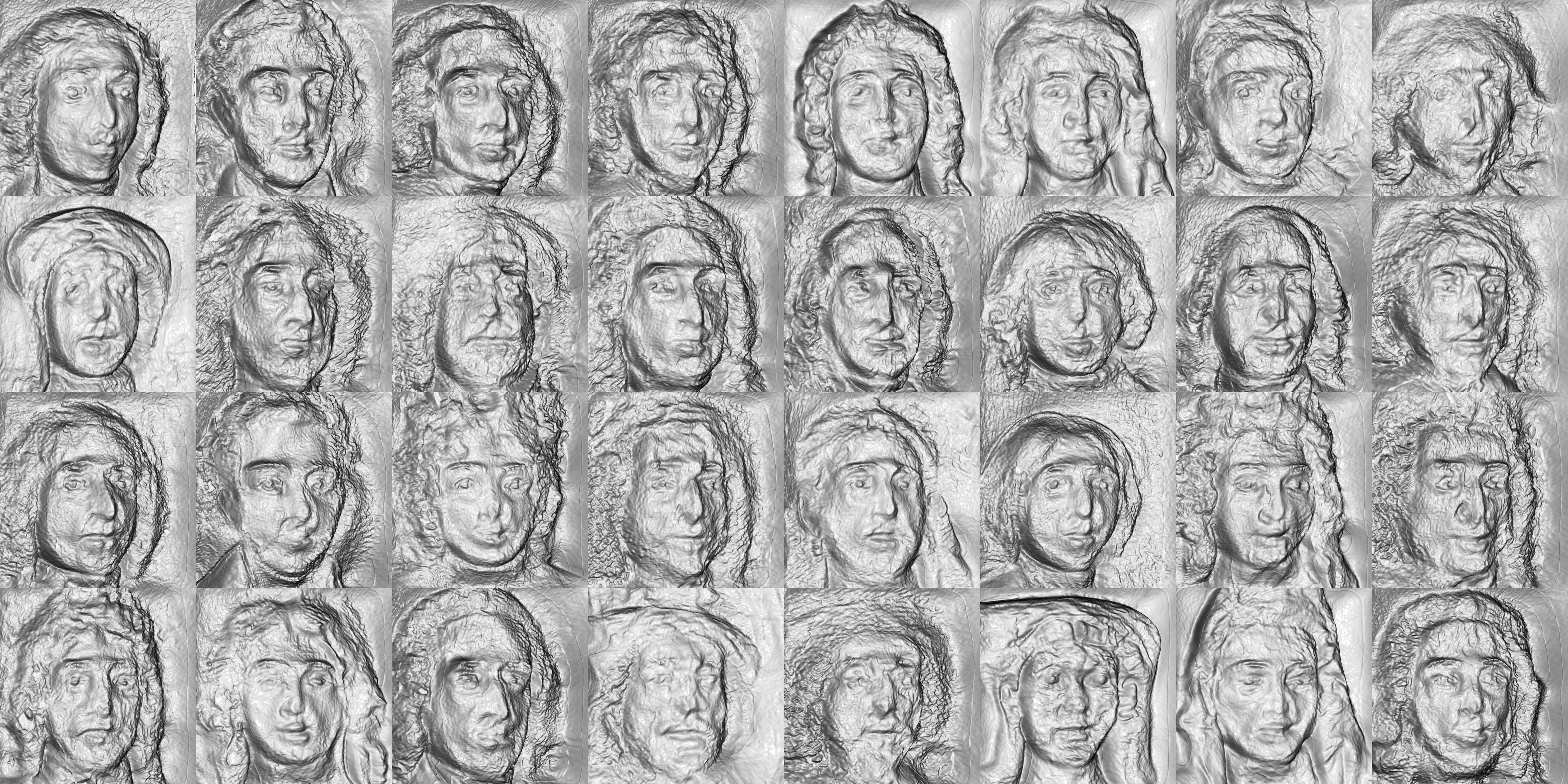}
        \captionsetup{width=\textwidth}
        \caption{Geometries for corresponding renderings in~\figref{supp fig: uncurated metfaces, rgb}.
        }
        \label{supp fig: uncurated metfaces, mesh}
    \end{subfigure}%
    \caption{
    \textbf{Uncurated results on MetFaces.} From top to bottom, left to right, we show generations with seed 1-32. Results are generated with truncation $\psi = 0.7$~\cite{Karras2019ASG}.
    }
    \label{supp fig: uncurated metfaces}
\end{figure}

We provide uncurated results on FFHQ (\figref{supp fig: uncurated ffhq}), \afhq~(\figref{supp fig: uncurated afhq}), and MetFaces (\figref{supp fig: uncurated metfaces}). We observe 
GMPI to generate high-quality geometry.

\subsection{Style Mixing}

\begin{figure}[!t]
    \centering
    \captionsetup[subfigure]{aboveskip=1pt}
    \begin{subfigure}{0.8\textwidth}
        \centering
        \includegraphics[width=\textwidth]{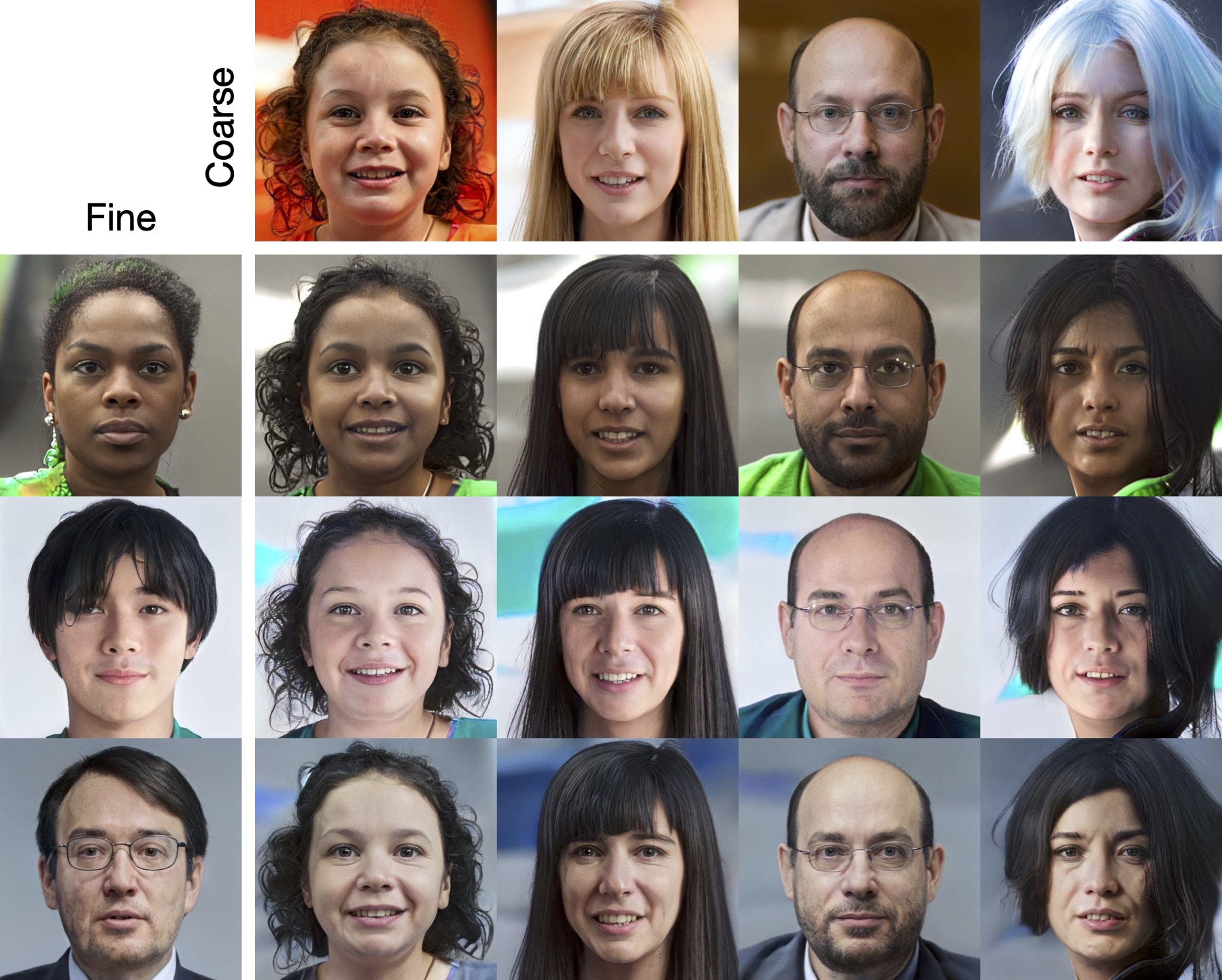}
        \captionsetup{width=\textwidth}
        \caption{Renderings for corresponding geometries in~\figref{supp fig: style mix ffhq512, mesh}.
        }
        \label{supp fig: style mix ffhq512, rgb}
    \end{subfigure}%
    \hfill
    \begin{subfigure}{0.8\textwidth}
        \centering
        \includegraphics[width=\textwidth]{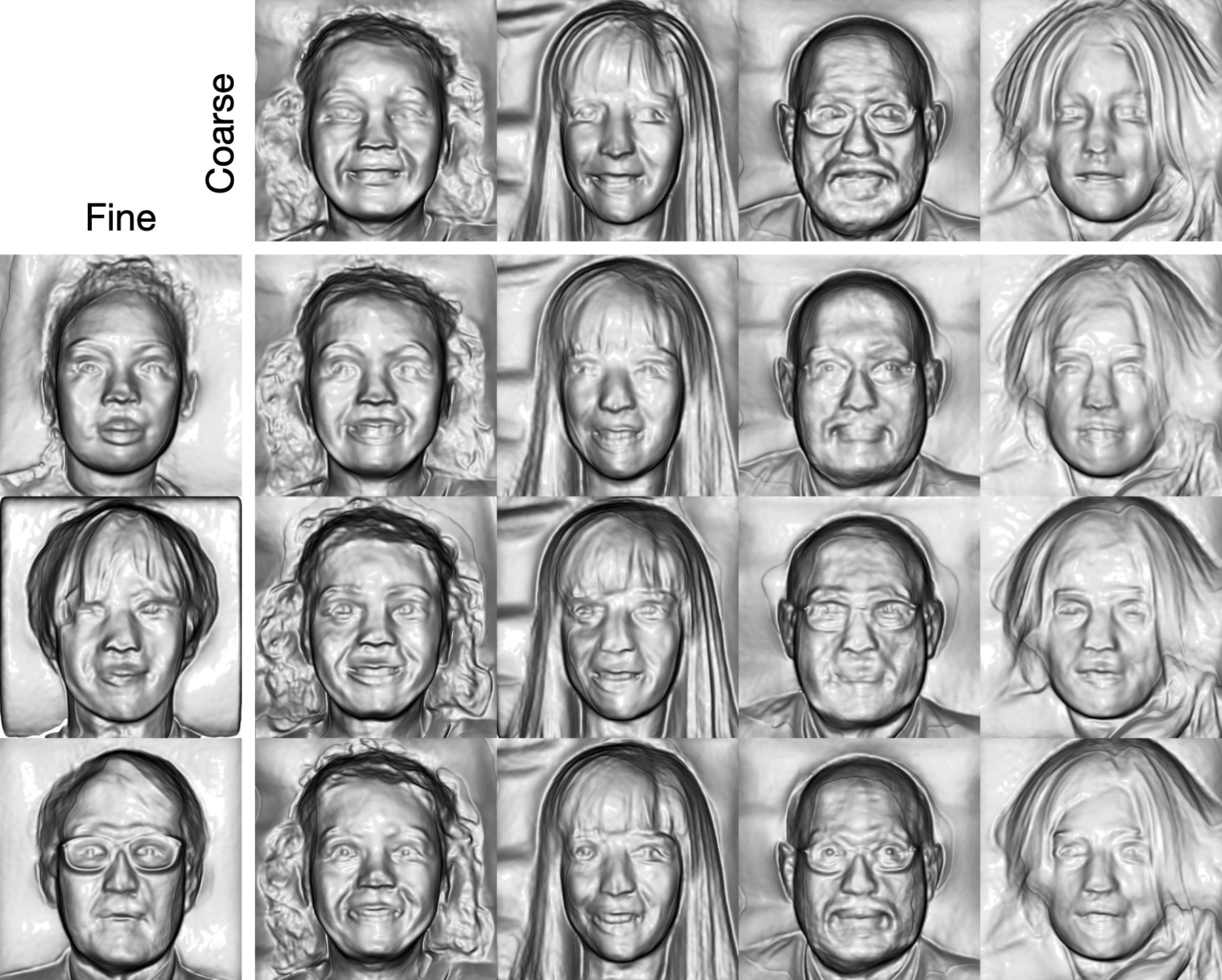}
        \captionsetup{width=\textwidth}
        \caption{Geometries for corresponding renderings in~\figref{supp fig: style mix ffhq512, rgb}.
        }
        \label{supp fig: style mix ffhq512, mesh}
    \end{subfigure}%
    \caption{
    \textbf{Style mixing on FFHQ.}
    Results don't use truncation,~\ie,~$\psi = 1.0$.
    To obtain each cell in the bottom right grid, we replace lower-level style embeddings $\bm{\omega}$ (\equref{eq: stylegan w}) in the \textit{Fine} column with the corresponding $\bm{\omega}$ from the \textit{Coarse} row.
    We observe,  GMPI enables semantic editing: lower-level $\bm{\omega}$ (layers 0 -- 6) control the shape while upper-level $\bm{\omega}$ (layers 7 and higher) determine fine-grained styles.
    }
    \label{supp fig: style mix ffhq512}
\end{figure}

\begin{figure}[!t]
    \centering
    \captionsetup[subfigure]{aboveskip=1pt}
    \begin{subfigure}{0.8\textwidth}
        \centering
        \includegraphics[width=\textwidth]{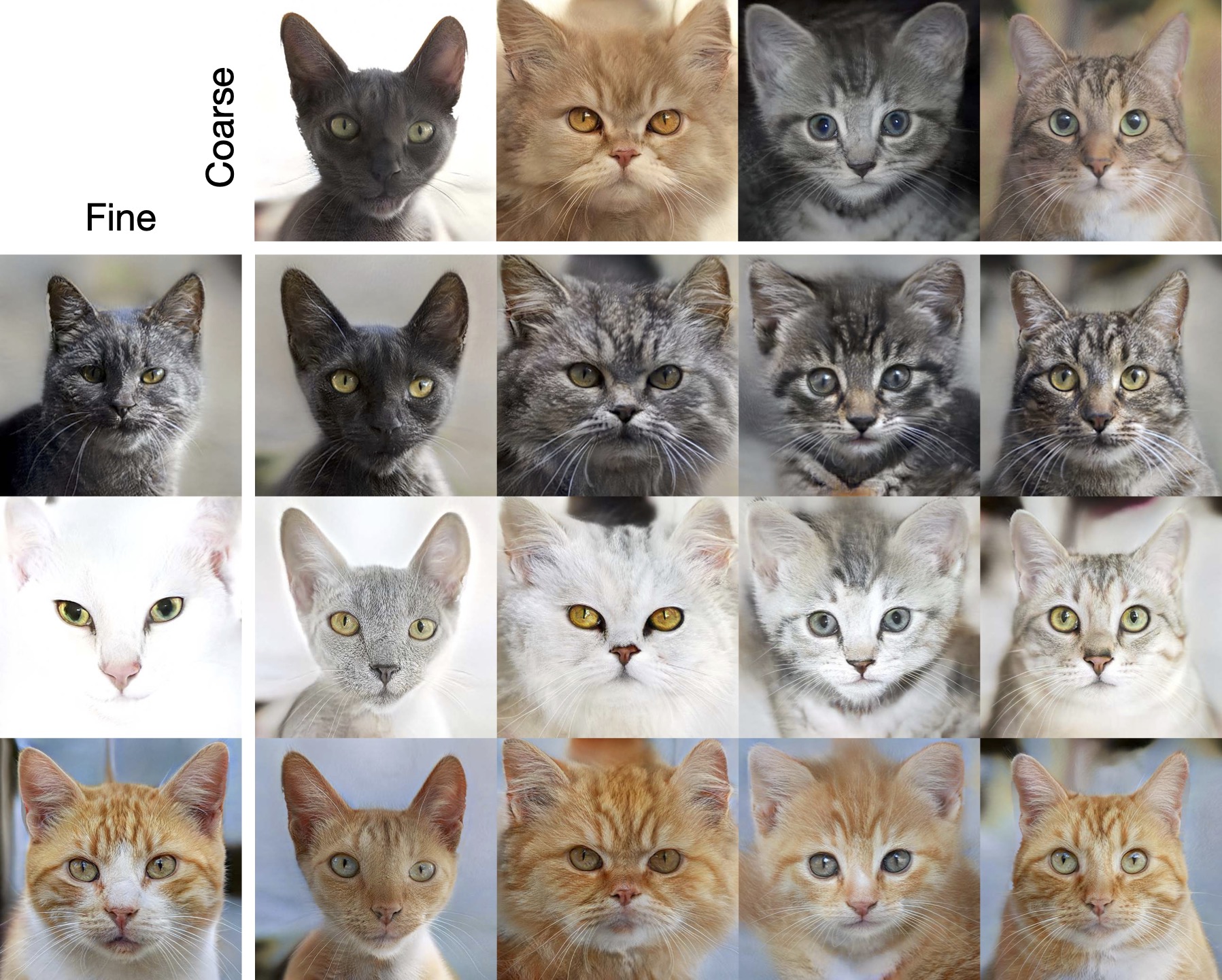}
        \captionsetup{width=\textwidth}
        \caption{Renderings for corresponding geometries in~\figref{supp fig: style mix afhq, mesh}.
        }
        \label{supp fig: style mix afhq, rgb}
    \end{subfigure}%
    \hfill
    \begin{subfigure}{0.8\textwidth}
        \centering
        \includegraphics[width=\textwidth]{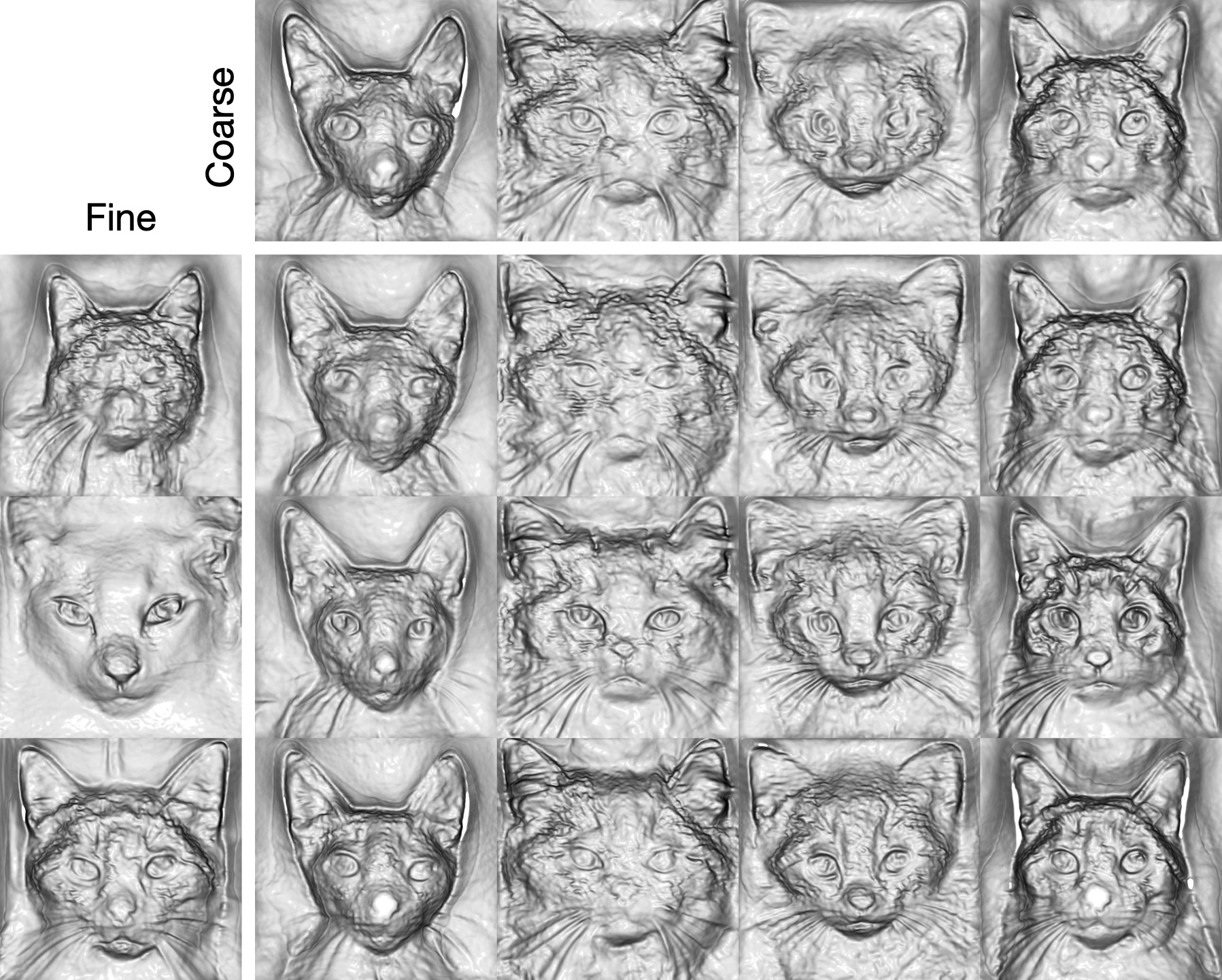}
        \captionsetup{width=\textwidth}
        \caption{Geometries for corresponding renderings in~\figref{supp fig: style mix afhq, rgb}.
        }
        \label{supp fig: style mix afhq, mesh}
    \end{subfigure}%
    \caption{
    \textbf{Style mixing on \afhq.}
    Results don't use truncation,~\ie,~$\psi = 1.0$.
    To obtain each cell in the bottom right grid, we replace lower-level style embeddings $\bm{\omega}$ (\equref{eq: stylegan w}) in the \textit{Fine} column with the corresponding $\bm{\omega}$ from the \textit{Coarse} row.
    We observe,  GMPI enables semantic editing: lower-level $\bm{\omega}$ (layers 0 -- 6) control the shape while upper-level $\bm{\omega}$ (layers 7 and higher) determine fine-grained styles.
    }
    \label{supp fig: style mix afhq}
\end{figure}

\begin{figure}[!t]
    \centering
    \captionsetup[subfigure]{aboveskip=1pt}
    \begin{subfigure}{0.8\textwidth}
        \centering
        \includegraphics[width=\textwidth]{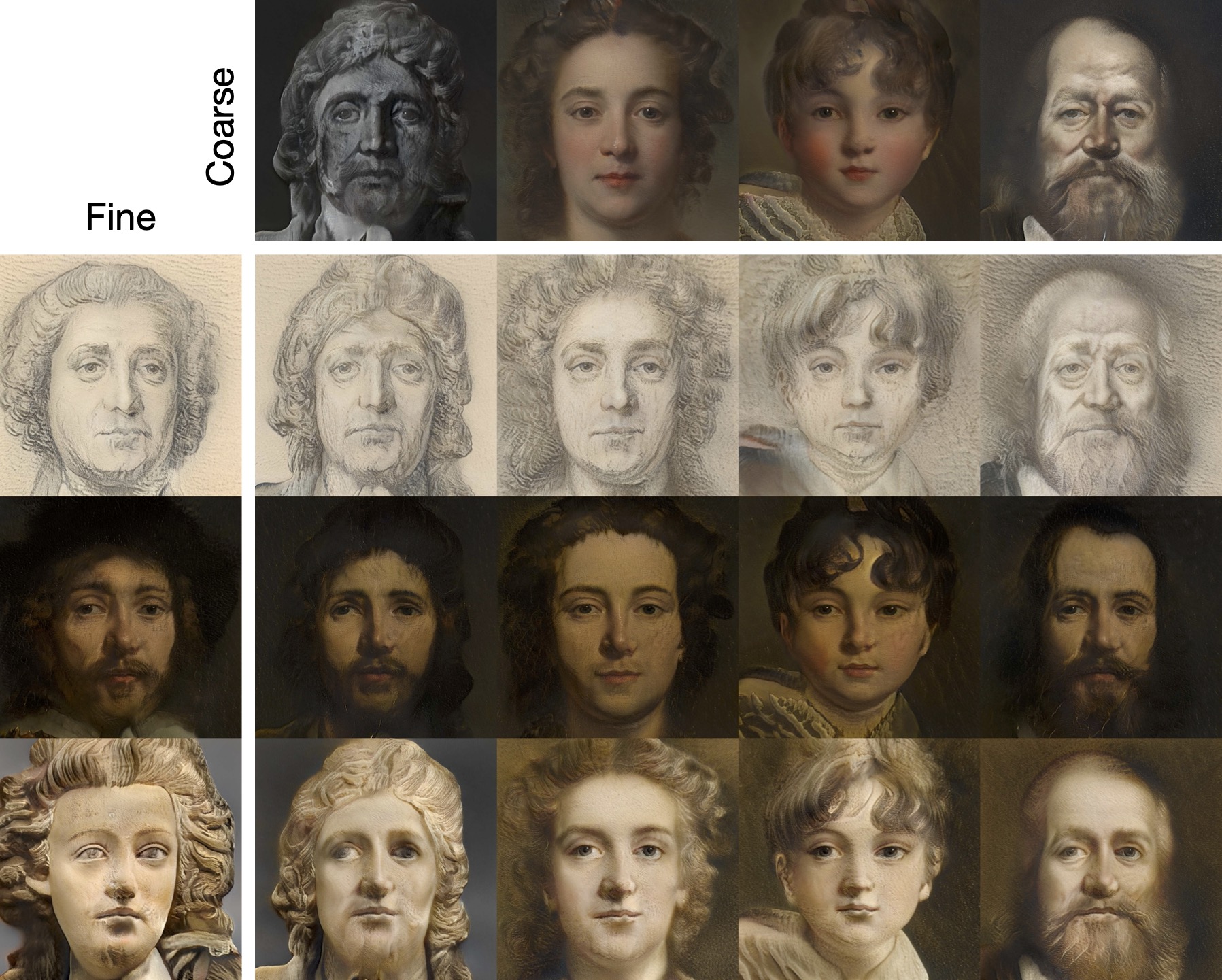}
        \captionsetup{width=\textwidth}
        \caption{Renderings for corresponding geometries in~\figref{supp fig: style mix metfaces, mesh}.
        }
        \label{supp fig: style mix metfaces, rgb}
    \end{subfigure}%
    \hfill
    \begin{subfigure}{0.8\textwidth}
        \centering
        \includegraphics[width=\textwidth]{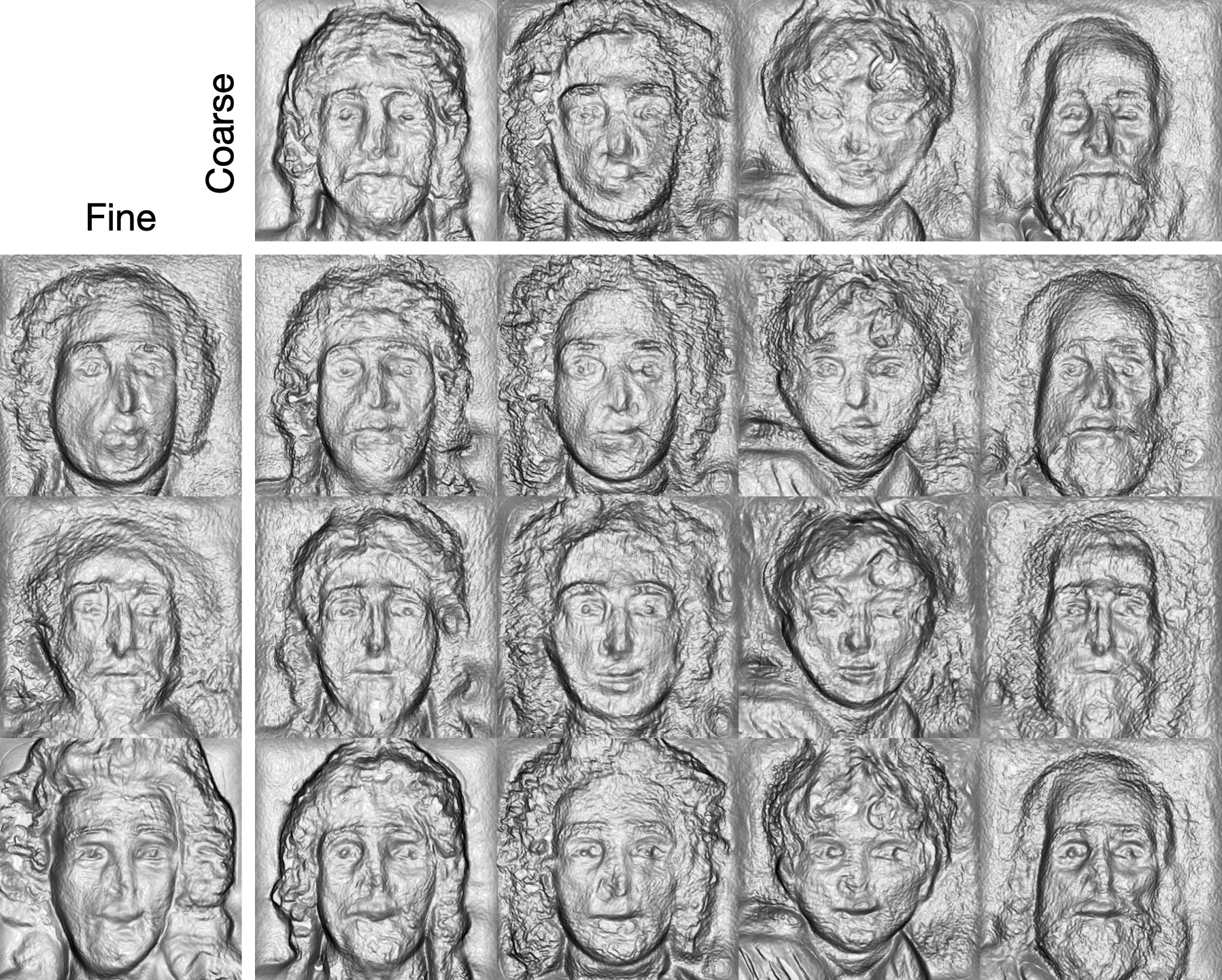}
        \captionsetup{width=\textwidth}
        \caption{Geometries for corresponding renderings in~\figref{supp fig: style mix metfaces, rgb}.
        }
        \label{supp fig: style mix metfaces, mesh}
    \end{subfigure}%
    \caption{
    \textbf{Style mixing on MetFaces.}
    Results don't use truncation,~\ie,~$\psi = 1.0$.
    To obtain each cell in the bottom right grid, we replace lower-level style embeddings $\bm{\omega}$ (\equref{eq: stylegan w}) in the \textit{Fine} column with the corresponding $\bm{\omega}$ from the \textit{Coarse} row.
    We observe,  GMPI enables semantic editing: lower-level $\bm{\omega}$ (layers 0 -- 6) control the shape while upper-level $\bm{\omega}$ (layers 7 and higher) determine fine-grained styles.
    }
    \label{supp fig: style mix metfaces}
\end{figure}

We illustrate style mixing~\cite{Karras2019ASG} results on FFHQ (\figref{supp fig: style mix ffhq512}), \afhq~(\figref{supp fig: style mix afhq}), and MetFaces (\figref{supp fig: style mix metfaces}).
GMPI successfully disentangles coarse and fine levels of generations.

\subsection{More Results}

This supplementary material  also includes an interactive viewer for the generated MPI representations and an HTML page with videos to illustrate  generations from GMPI.

\end{document}